%% file: main.tex
\documentclass[10pt]{article}

\usepackage[preprint]{tmlr}

\usepackage{amsmath,amssymb}
\usepackage{graphicx}
\usepackage{booktabs}
\usepackage{subcaption}
\usepackage{multirow}
\usepackage[table]{xcolor}
\usepackage{array}
\usepackage{needspace}
\usepackage{placeins}
\usepackage{algorithm}
\usepackage{algpseudocode}
\usepackage{paperabbrv}

\algrenewcommand\algorithmicrequire{\textbf{Input:}}
\algrenewcommand\algorithmicensure{\textbf{Output:}}
\makeatletter
\algrenewcommand\ALG@beginalgorithmic{\small}
\makeatother

\definecolor{journalblue}{RGB}{0,86,125}
\usepackage{hyperref}
\hypersetup{
  breaklinks=true,
  colorlinks=true,
  linkcolor=journalblue,
  urlcolor=journalblue,
  citecolor=journalblue
}

\let\cite\citep
\newcommand{\tabcite}[1]{~\citep{#1}}
\newcommand{\best}[1]{\textbf{#1}}
\newcommand{\HM}{\ensuremath{\mathrm{HM}}}
\newcommand{\second}[1]{\underline{#1}}

\begin{document}

\title{ProCAP: Probabilistic Cross-Attentive Prompt Learning for Vision--Language Models%
\thanks{Generative AI tools were used to assist with LaTeX formatting, language revision, and consistency checks of literature tables. The authors verified all citations, numerical results, scientific claims, and final content and take full responsibility for the manuscript.}}

\author{%
\name Hiwa Azeez Abbas$^{1}$ \quad Fatemeh Daneshfar$^{2}$ \quad Moloud Abdar$^{3}$ \\
\addr $^{1}$University of Kurdistan, Sanandaj, Iran \\
\addr $^{2}$Department of Computer Engineering, University of Kurdistan, Sanandaj, Iran \\
\addr $^{3}$CHIRP, Child Health Research Centre, The University of Queensland, Brisbane, Australia
}

\def\month{September}
\def\year{2026}
\def\openreview{\url{https://openreview.net/forum?id=XXXX}}

\maketitle

\begin{abstract}
Pre-trained vision--language models such as CLIP can recognize new categories via prompting, but they often struggle when labeled data are scarce or the test distribution shifts. Prompt learning adapts only a small set of parameters while keeping the backbone frozen, yet many existing multimodal prompt learners couple the visual and textual branches weakly and can be brittle in low-shot regimes. We propose \textbf{ProCAP}, a probabilistic cross-attentive prompt learning framework that improves cross-modal interaction and training stability without updating any CLIP weights: it learns both visual and textual prompt tokens and links them through \emph{stacked bidirectional} multi-head cross-attention so the two branches refine each other across prompt depth. To reduce overfitting under limited supervision, we parameterize prompt tokens with Gaussian means and variances and regularize them with lightweight KL and $L_2$ penalties, and we further add a compact symmetric InfoNCE head that aligns cross-attended image features with class-level text representations in a shared low-dimensional space. Across few-shot base-to-novel generalization on 11 datasets, cross-dataset transfer, and domain generalization on ImageNet shift benchmarks, ProCAP achieves strong aggregate base-to-novel performance and competitive transfer performance while keeping the CLIP backbone unchanged.
\end{abstract}

\noindent\textbf{Keywords:} CLIP; multimodal prompt learning; cross-attention; prompt regularization; few-shot learning.

\noindent\textbf{Code:} \url{https://github.com/hiwea/ProCAP}

\section{Introduction}
\label{sec:intro}

Vision--language models (VLMs) such as CLIP~\cite{radford2021clip} learn a shared image--text embedding space from large-scale contrastive data and enable strong zero-shot recognition with natural language prompts. In practice, however, downstream datasets rarely match CLIP’s pretraining distribution: domains shift, class vocabularies change, and supervision is often limited to a few labeled examples per class. Naively fine-tuning the whole backbone is computationally costly and can destroy CLIP’s transferable knowledge. Prompt learning has therefore become a standard adaptation strategy: it keeps CLIP frozen and learns only a small set of context tokens, achieving efficient transfer under low-shot supervision~\cite{du2024ipo}. Different prompt-tuning objectives further improve stability and transfer under limited supervision~\cite{roy2024coprompt}.

Recent work has moved from text-only prompts to multimodal prompting, where both text and image branches are adapted for stronger transfer~\cite{khattak2023maple,wu2023ammpl,kim2024aapl}. For example, AMMPL~\cite{wu2023ammpl} adjusts multimodal prompts with adaptive weighting across modalities, while AAPL~\cite{kim2024aapl} focuses on attribute-aware prompts for fine-grained control. Style-Pro~\cite{alipour2024stylepro} explicitly encourages robustness to style shift through style-guided prompting. Beyond these designs, structured multimodal architectures go further: MUAP~\cite{dai2024muap}, DPC~\cite{li2025dpc}, VAMP~\cite{cheng2025vamp}, DiMPLe~\cite{rahman2025dimple} and MuGCP~\cite{yang2025mugcp} strengthen image--text interaction within CLIP-style frameworks. In-context prompting further shows that large vision--language models can adapt to new tasks with only a few exemplars provided at inference~\cite{yin2024incpl}.

Despite this progress, two challenges remain for MaPLe-style multimodal prompting in few-shot and base-to-novel transfer. First, cross-modal interaction is still not tight enough: even multimodal prompt learners can behave like two loosely coupled modules, and shallow or one-directional links limit how much visual evidence can refine textual semantics during adaptation~\cite{khattak2023maple}. Dual-prompt designs partially address this but still leave room for richer bidirectional interaction between the image and text branches~\cite{li2025dpc}. Second, optimizing prompts reliably in low-shot settings is difficult. With very limited supervision, prompt parameters may latch onto incidental cues and generalize poorly to new domains or novel classes, which motivates methods that explicitly improve stability and robustness during adaptation~\cite{mistretta2024kdpl,alipour2024stylepro,zhang2024dept}.

\begin{figure}[!t]
\centering
\captionsetup{skip=7pt}
\captionsetup[subfigure]{font=footnotesize,labelfont=bf,textfont=normalfont,skip=1pt}
\newcommand{\panelH}{0.24\textheight}

\begin{subfigure}[t]{\textwidth}
  \centering
  \includegraphics[width=0.9\linewidth,height=\panelH,keepaspectratio]{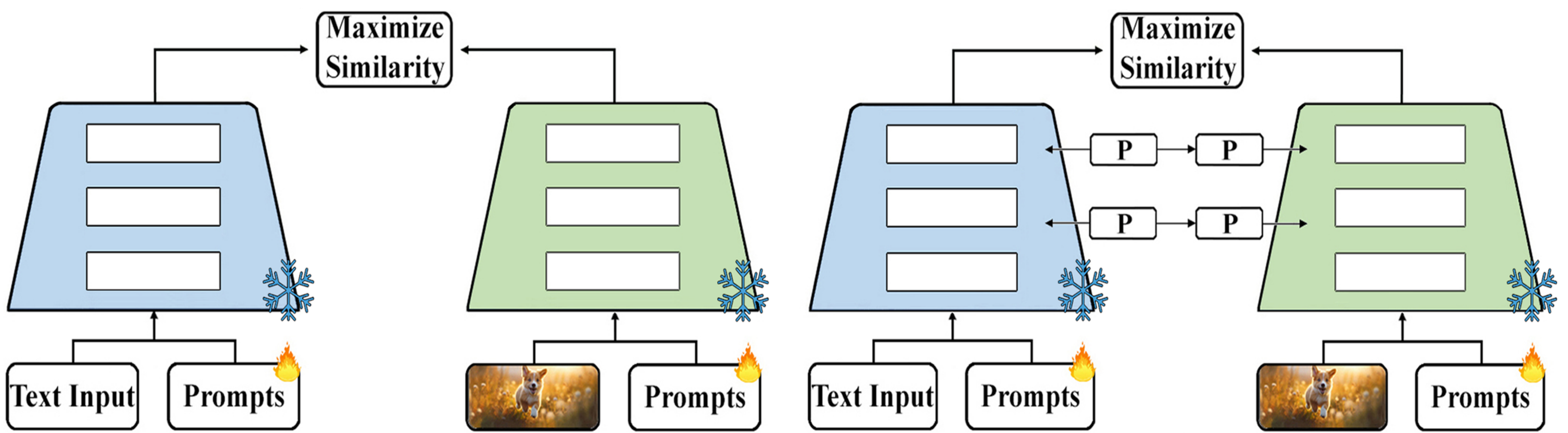}
  \subcaption{Multi-template prompting and MaPLe-style prompting.}
  \label{fig:prompting_and_radar_a}
\end{subfigure}

\vspace{2mm}

\begin{subfigure}[t]{\textwidth}
  \centering
  \includegraphics[width=0.9\linewidth,height=\panelH,keepaspectratio]{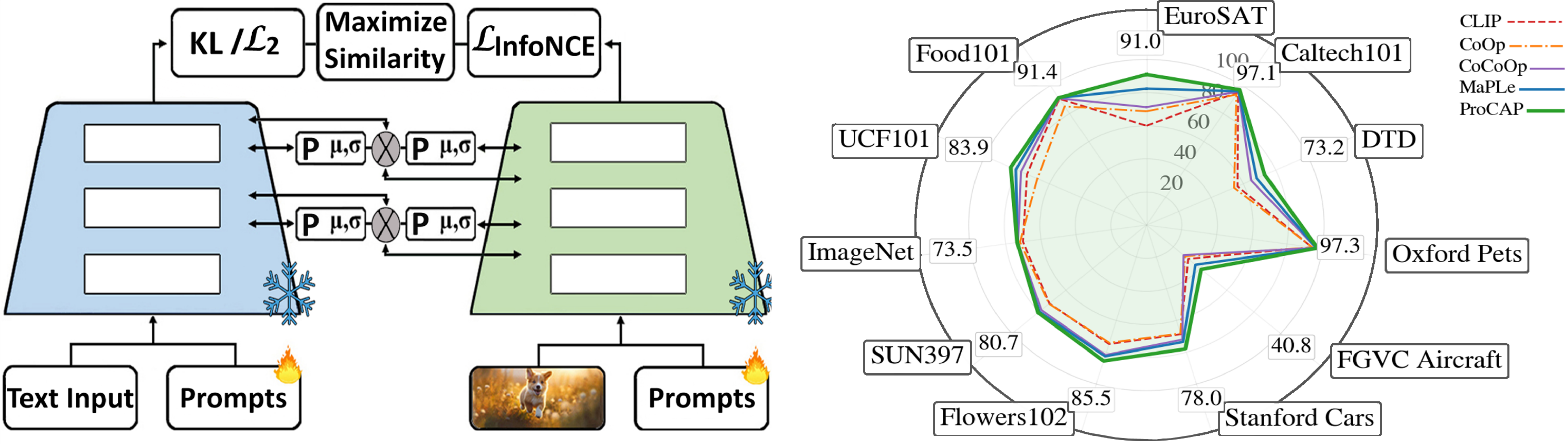}
  \subcaption{ProCAP prompting and base-to-novel \HM{} across benchmarks.}
  \label{fig:prompting_and_radar_b}
\end{subfigure}

\vspace{1mm}

\caption{Overview of prompt learning designs and their effect on base-to-novel generalization.
(a) Multi-template prompting and MaPLe-style prompting.
(b) ProCAP prompting and base-to-novel \HM{} across benchmarks.}
\label{fig:prompting_and_radar}
\end{figure}

In this work, we tackle both issues with ProCAP (\textbf{Pro}babilistic \textbf{C}ross-\textbf{A}ttentive \textbf{P}rompt learning), a strengthened multimodal prompt learner that keeps CLIP fully frozen while improving how modalities interact and how prompts are optimized (see Figure~\ref{fig:prompting_and_radar}). Our design builds directly on MaPLe-style multimodal prompts but introduces a more tightly coupled interaction mechanism. Specifically, we introduce learnable visual prompts alongside text prompts, project image-side prompts into the text embedding space, and enable stacked bidirectional multi-head cross-attention between deep visual and textual prompt embeddings. This creates a direct two-way refinement pathway in which visual prompts attend to text prompts and text prompts attend to visual prompts, allowing each modality to guide the other throughout training and tightening cross-modal coupling beyond prior multimodal prompting architectures. Figure~\ref{fig:prompting_and_radar_b} additionally provides an intuitive cross-dataset view of base-to-novel generalization using \HM{}, where \HM{} balances performance on base and novel classes; compared to CoOp/CoCoOp and MaPLe, ProCAP improves this balance on several challenging benchmarks. Our stacked formulation is inspired by hierarchical multimodal prompting but places cross-attention at the center of the interaction instead of treating it as a peripheral component.

To stabilize low-shot adaptation, we augment the standard prompt-learning classification objective with a lightweight symmetric InfoNCE head. Using cross-attended image features and class-level text representations (the final CLIP text feature produced from each class name together with its learned textual prompts), we enforce in a shared low-dimensional space that images are close to the representation of their own class and far from those of other classes, with a learnable temperature for similarity scaling. Finally, we incorporate Gaussian prompt-space regularization by modeling each prompt token with a learned mean and variance, and applying KL and $L_2$ penalties only to these prompt parameters. This reduces drift and overfitting while leaving all CLIP backbone weights frozen and makes prompt optimization more reliable under domain and class shift.

Taken together, these components yield a simple, single-stage training recipe that updates only prompts, cross-attention modules and the additional InfoNCE head. Across standard few-shot and base-to-novel benchmarks, ProCAP consistently improves over MaPLe-style multimodal prompting while preserving CLIP’s transferable representations~\cite{khattak2023maple}. We also discuss recent prompt-learning approaches such as CoPrompt~\cite{roy2024coprompt}, CasPL~\cite{wu2024caspl}, MoCoOp~\cite{du2024mocoop}, and DPC~\cite{li2025dpc}.

\medskip
\noindent Our primary contributions are summarized as follows.

{%
\renewcommand{\labelitemi}{$\bullet$}%
\begin{itemize}
    \item We propose ProCAP, a probabilistic cross-attentive multimodal prompt learner that couples text and image prompts through stacked bidirectional multi-head cross-attention across prompt depth, giving each modality a direct pathway to refine the other under a frozen CLIP backbone.
    \item We introduce a lightweight symmetric InfoNCE head over cross-attended image features and class-level text prompt representations, tightening image--text alignment beyond the standard prompt-learning classification objective in low-shot settings.
    \item We add Gaussian prompt-space regularization with KL/$L_2$ penalties, which stabilizes few-shot optimization and reduces prompt overfitting while leaving all CLIP backbone weights frozen.
\item We provide a simple end-to-end training recipe on top of frozen CLIP and, through extensive experiments on 11 standard image-classification benchmarks, show strong aggregate performance relative to MaPLe-style multimodal prompting and recent prompt-learning baselines.
\end{itemize}
}%

\section{Related Work}
\label{sec:related}

\noindent\textbf{Vision--Language Models.}
CLIP~\cite{radford2021clip} showed that contrastive training on large-scale image--text pairs can produce strong, reusable vision--language representations, making lightweight adaptation attractive for downstream recognition tasks. Many follow-up methods build on this idea by changing how prompts, regularization, or cross-modal interaction are designed. MaPLe~\cite{khattak2023maple} introduces multimodal prompts to better couple visual and textual features. CoPrompt~\cite{roy2024coprompt} adds consistency constraints to protect base-class performance, while Style-Pro~\cite{alipour2024stylepro} focuses on robustness to style shifts. KDPL~\cite{mistretta2024kdpl} learns prompts from unlabeled data to improve zero-shot generalization, and methods such as DPC~\cite{li2025dpc}, VAMP~\cite{cheng2025vamp}, DiMPLe~\cite{rahman2025dimple}, and MuGCP~\cite{yang2025mugcp} further explore structured or adaptive prompting mechanisms for stronger image--text alignment.

\medskip
\noindent\textbf{Prompt Learning in Vision--Language Models.}
Prompt learning adapts pre-trained models by optimizing a small set of context embeddings instead of updating all weights, which reduces computation and helps avoid catastrophic forgetting. Early work mainly studied text-only prompts, but multimodal variants have proved more effective in practice. AMMPL~\cite{wu2023ammpl}, AAPL~\cite{kim2024aapl}, and IPO~\cite{du2024ipo} extend prompts with visual cues, attribute structure, and language-model guidance. MoCoOp~\cite{du2024mocoop} and CasPL~\cite{wu2024caspl} further specialize prompts across inputs and datasets. Across these methods, keeping a CLIP-style contrastive objective~\cite{radford2021clip} while adding task-specific prompt modules has emerged as a robust adaptation recipe.

\medskip
\noindent\textbf{Relation to Recent Prompt-Learning Methods.}
ProCAP is related to VAMP and DPC, but differs in setting and mechanism. VAMP uses sample-specific variational prompting with inference-time sampling, while DPC focuses on dual-prompt collaboration to decouple optimization for base and new classes. In contrast, ProCAP keeps the standard frozen-CLIP prompt-learning setting and jointly optimizes bidirectional visual--text prompt interaction, Gaussian prompt-space regularization, and symmetric image--class alignment without external descriptions, test-time sampling, or a teacher/plugin framework. Thus, ProCAP should be viewed as a unified prompt-learning framework rather than a direct reuse of any single prior component.

\medskip
\noindent\textbf{Recent Prompt-Learning Developments.}
Several recent studies further develop multimodal adaptation for CLIP. HI$^{2}$MA combines hierarchical intra- and inter-modal adapters with knowledge reinforcement and bidirectional cross-modal attention~\cite{dong2026hi2ma}. DCPL constructs diversified composite prompts from shared roots and modality-specific branches~\cite{mei2026dcpl}, whereas MPCLe forms input-adaptive prompts by selecting tokens from learned visual and textual codebooks~\cite{zhang2026mpcle}. MMLoP instead represents deep multimodal prompts through low-rank factors and supplements them with consistency and drift-correction objectives~\cite{ghiasvand2026mmlop}. ProCAP is complementary to these approaches: its central mechanism is stacked bidirectional cross-attention between probabilistic visual and textual prompts, optimized jointly with prompt-space regularization and symmetric image--class contrastive alignment.

\section{Method}
\label{sec:method}

\begin{figure}
    \centering
\includegraphics[width=\columnwidth]{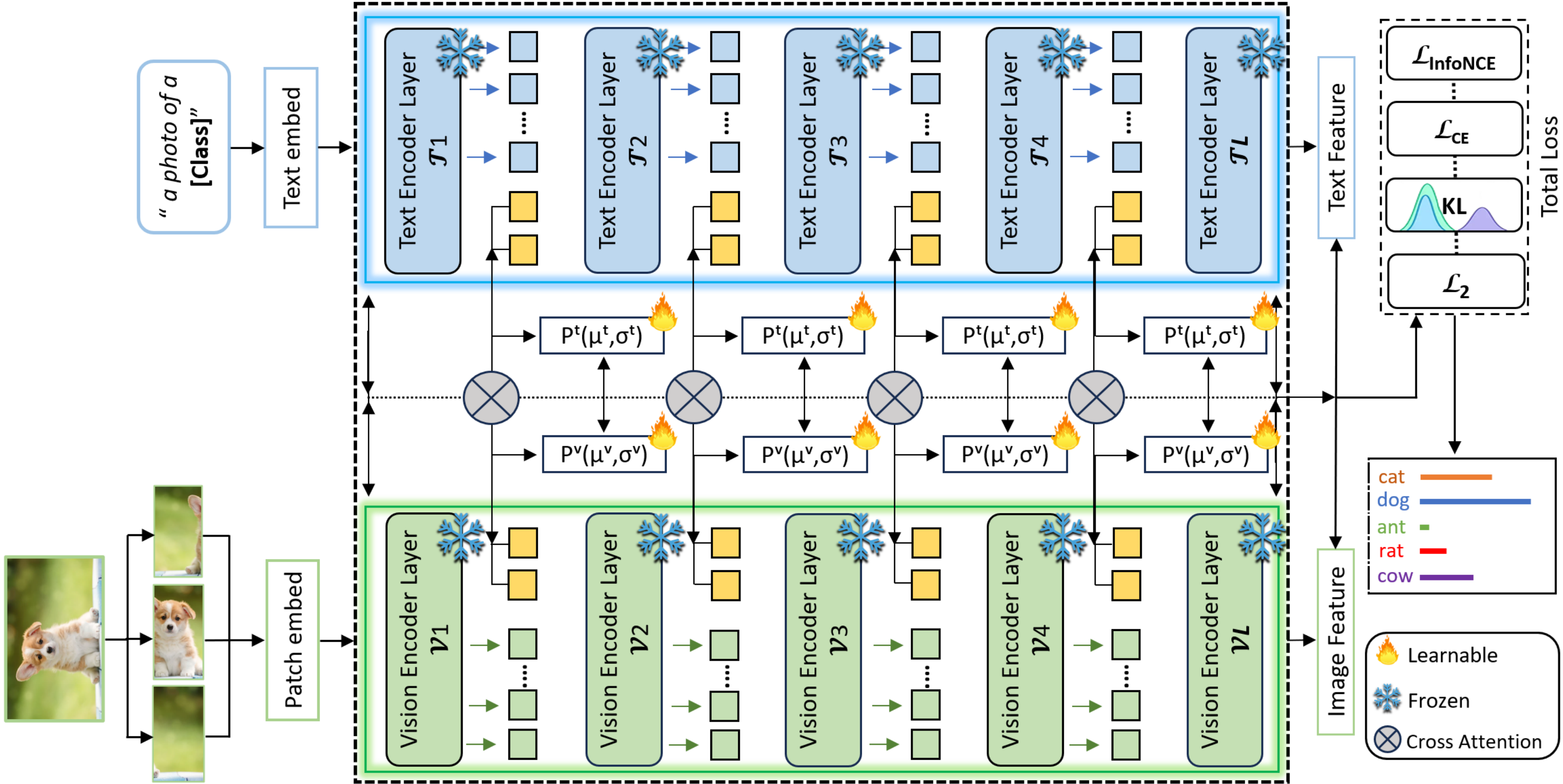}
    \caption{Overview of the proposed ProCAP framework. ProCAP builds on a frozen CLIP backbone and introduces cross-attentive multimodal prompts, probabilistic (Gaussian) prompt parameterization, and a symmetric InfoNCE head for improved few-shot and base-to-novel transfer.}
    \label{fig:framework}
\end{figure}

In this section we present ProCAP, a probabilistic cross-attentive prompt learner built on top of a frozen CLIP vision--language backbone. Instead of fine-tuning CLIP, ProCAP only modifies the prompt space and a few lightweight heads (see Figure~\ref{fig:framework}). First, we introduce a cross-attentive multimodal prompt module that couples image and text prompts through stacked bidirectional multi-head cross-attention across prompt depth. Second, we model each prompt token with Gaussian parameters and regularize them with KL and $L_2$ terms, using this as prompt-space regularization to reduce overfitting in few-shot and base-to-novel settings. Third, we add a symmetric InfoNCE head over cross-attended image features and class-level text representations to explicitly tighten image--class alignment beyond the standard prompt-learning classification objective. We first review the CLIP and MaPLe-style multimodal prompting setup that we build on, and then detail our cross-attentive Gaussian prompts, symmetric InfoNCE formulation, and overall training loss.

\subsection{Preliminaries: CLIP and Multimodal Prompting}
\label{subsec:prelim}

We briefly review the CLIP and MaPLe-style multimodal prompting setup that ProCAP builds on. Let $f_{\mathrm{img}}$ and $f_{\mathrm{text}}$ denote the frozen CLIP image and text encoders, respectively. Given an image $x$ and a class label $y$ with textual name $c_y$ (e.g., \textit{``dog''}), CLIP forms a prompt by concatenating learnable or hand-crafted context tokens with the class name, encodes it with $f_{\mathrm{text}}$, and maximizes the similarity between the resulting text feature and the corresponding image feature.

During supervised prompt learning, the class text features act as classifier weights and the learnable prompts are optimized with cross-entropy over image-to-class logits:
\begin{equation}
\label{eq:clip}
p(y{=}k\mid x_i)
= \frac{\exp(\mathrm{sim}(v_i,t_k)/\tau)}{\sum_{c=1}^{K}\exp(\mathrm{sim}(v_i,t_c)/\tau)},
\qquad
\mathcal{L}_{\mathrm{CE}}
= -\frac{1}{N}\sum_{i=1}^{N}\log p(y_i\mid x_i),
\end{equation}
where $v_i=f_{\mathrm{img}}(x_i)$, $t_k=f_{\mathrm{text}}(\mathrm{prompt}(c_k))$, $\mathrm{sim}(\cdot,\cdot)$ is cosine similarity, $\tau$ is a learnable temperature, $K$ is the number of classes, and $N$ is the batch size.

Prompt learning keeps $f_{\mathrm{img}}$ and $f_{\mathrm{text}}$ frozen and only optimizes a small set of context embeddings, often called \emph{prompt tokens}. MaPLe-style multimodal prompting~\cite{khattak2023maple} further introduces visual prompts injected into the image encoder together with text prompts injected into the text encoder, and couples them through a shared projection space. This improves alignment between modalities while preserving CLIP's transferable representations.

\subsection{ProCAP: Cross-Attentive Multimodal Prompts}
\label{subsec:procapprompts}

Our first contribution is a cross-attentive multimodal prompt module that couples image and text prompts more tightly than MaPLe while keeping the CLIP backbone frozen.

\paragraph{Multimodal prompt tokens.}
Following MaPLe, we learn both visual and textual prompts. We write
$\mathbf{P}^{v} \in \mathbb{R}^{L_p \times d_v}$ for the visual prompts and
$\mathbf{P}^{t} \in \mathbb{R}^{L_p \times d_t}$ for the text prompts, where $L_p$ is the number of prompt tokens and $d_v$ and $d_t$ are the native hidden widths of the frozen vision and text transformers, respectively. In the vision encoder, we prepend $\mathbf{P}^{v}$ to the patch-token sequence at selected transformer layers (prompt depth). In the text encoder, we prepend $\mathbf{P}^{t}$ to the tokenized class prompt (e.g., \textit{``a photo of a [Class]''}), consistent with the notation in Figure~\ref{fig:framework}.

\paragraph{Shared prompt space and bidirectional cross-attention.}
To allow direct interaction between modalities with different native widths, we use a learnable visual-to-text projection
$W_p:\mathbb{R}^{d_v}\!\rightarrow\!\mathbb{R}^{d_t}$ and define
$\widetilde{\mathbf{P}}^{v}_{\ell}=W_p\mathbf{P}^{v}_{\ell}$.
At layer $\ell$, the visual-to-text and text-to-visual directions use separate multi-head cross-attention parameters. We denote the visual update abstractly as
$\mathrm{MHA}_v(\mathbf{P}^{v}_{\ell},\mathbf{P}^{t}_{\ell},\mathbf{P}^{t}_{\ell})$, whose query/key/value projections map the native visual/text widths to the attention-head space and whose output projection returns to $d_v$. The reverse update uses textual queries with the projected visual prompts,
$\mathrm{MHA}_t(\mathbf{P}^{t}_{\ell},\widetilde{\mathbf{P}}^{v}_{\ell},\widetilde{\mathbf{P}}^{v}_{\ell})$, and returns to $d_t$. Each attention output is fused back into its current prompt stream through a residual update with normalization, so cross-modal information augments rather than replaces the existing prompt representation. Stacking these blocks across prompt depth yields a hierarchical bidirectional interaction pathway in which the two prompt streams are repeatedly refined while the CLIP transformer weights remain frozen.

\subsection{Gaussian Prompt Regularization}
\label{subsec:uncertainty}

Our second contribution is to parameterize prompt tokens with Gaussian means and variances, using them as prompt-space regularization in low-data regimes rather than as full Bayesian predictive uncertainty.

\paragraph{Gaussian prompt tokens.}
For each prompt token in modality $m\in\{v,t\}$, we maintain parameters
$(\boldsymbol{\mu}^{m}, \boldsymbol{\sigma}^{m}) \in \mathbb{R}^{d_m} \times \mathbb{R}^{d_m}$ and sample the actual prompt used during training via a standard reparameterization. For readability, we omit the modality superscript below:
\begin{equation}
\label{eq:reparameterization}
\mathbf{P}_j = \boldsymbol{\mu}_j + \boldsymbol{\sigma}_j \odot \boldsymbol{\epsilon}_j,
\qquad
\boldsymbol{\epsilon}_j \sim \mathcal{N}(\mathbf{0}, \mathbf{I}),
\end{equation}
where $\odot$ denotes element-wise multiplication and $j$ indexes a specific prompt token. In practice we parameterize $\log \boldsymbol{\sigma}^2$ for numerical stability. At inference time, we simply use the means $\boldsymbol{\mu}_j$ without sampling, so the model behaves deterministically.

\paragraph{KL and $L_2$ regularization.}
We regularize the distribution of prompts towards a zero-mean isotropic Gaussian prior through a KL divergence term:
\begin{equation}
\label{eq:kl}
\mathcal{L}_{\mathrm{KL}}
= \sum_{j} \mathrm{KL}\big(
\mathcal{N}(\boldsymbol{\mu}_j, \mathrm{diag}(\boldsymbol{\sigma}_j^2))
\;\|\;
\mathcal{N}(\mathbf{0}, \mathbf{I})
\big),
\end{equation}
where the sum runs over all prompt tokens in both modalities. We further add an $L_2$ penalty on the means to discourage overly large prompt magnitudes:
\begin{equation}
\label{eq:l2}
\mathcal{L}_{2} = \sum_{j} \|\boldsymbol{\mu}_j\|_2^2.
\end{equation}
These regularizers act directly on the prompt space and encourage prompts to stay close to a simple prior unless strongly supported by the data, which improves robustness in few-shot and base-to-novel transfer.

\subsection{Symmetric InfoNCE Head over Image and Class Representations}
\label{subsec:infonce}

Our third contribution is a lightweight symmetric InfoNCE head that aligns cross-attended image features with class-level text representations in a low-dimensional space, complementing the standard prompt-learning classification objective.

\paragraph{Class-level text representations.}
For each class $k$, we obtain the final text feature $t_k$ by feeding the class name and its associated textual prompts through the text encoder. We then apply a small projection head $g_{\mathrm{text}}$ (e.g., an MLP with one hidden layer) to obtain a class-level representation:
\begin{equation}
\mathbf{c}_k = g_{\mathrm{text}}(t_k) \in \mathbb{R}^{d_h},
\end{equation}
where $d_h$ is a low-dimensional hidden size.

\paragraph{Image features for InfoNCE.}
For an input image $x_i$, we take the output $v_i$ of the image encoder with visual prompts and cross-attention, and project it into the same space using $g_{\mathrm{img}}$:
\begin{equation}
\mathbf{z}_i = g_{\mathrm{img}}(v_i) \in \mathbb{R}^{d_h}.
\end{equation}
We $\ell_2$-normalize both $\mathbf{z}_i$ and $\mathbf{c}_k$ before computing similarities.

\paragraph{Symmetric InfoNCE loss.}
Given a batch of image features $\{\mathbf{z}_i\}_{i=1}^N$ and labels $\{y_i\}_{i=1}^N$, together with class representations $\{\mathbf{c}_k\}_{k=1}^K$, let $\mathcal{P}_k=\{i:y_i=k\}$ and let $\mathcal{Y}_{B}$ denote the classes represented in the current batch. We use an image-to-class term
\begin{equation}
\mathcal{L}_{I\rightarrow C}
= -\frac{1}{N}\sum_{i=1}^{N}
\log\frac{\exp(\mathbf{z}_i^\top\mathbf{c}_{y_i}/\tau_{\mathrm{p}})}
{\sum_{k=1}^{K}\exp(\mathbf{z}_i^\top\mathbf{c}_{k}/\tau_{\mathrm{p}})},
\end{equation}
and a class-to-image multi-positive term
\begin{equation}
\mathcal{L}_{C\rightarrow I}
= -\frac{1}{|\mathcal{Y}_{B}|}\sum_{k\in\mathcal{Y}_{B}}
\log\frac{\sum_{i\in\mathcal{P}_k}\exp(\mathbf{c}_k^\top\mathbf{z}_i/\tau_{\mathrm{p}})}
{\sum_{j=1}^{N}\exp(\mathbf{c}_k^\top\mathbf{z}_j/\tau_{\mathrm{p}})}.
\end{equation}
The symmetric objective is
\begin{equation}
\label{eq:infonce}
\mathcal{L}_{\mathrm{InfoNCE}}
=\tfrac{1}{2}\bigl(\mathcal{L}_{I\rightarrow C}+\mathcal{L}_{C\rightarrow I}\bigr),
\end{equation}
where $\tau_{\mathrm{p}}$ is a learnable temperature separate from the CLIP temperature. The image-to-class direction contrasts each image against all class representations. In the reverse direction, every image in the batch that shares class $k$ is treated as a positive for $\mathbf{c}_k$; same-class images are therefore not used as false negatives when a batch contains multiple examples from the same class.

\begin{algorithm}[H]
\caption{ProCAP Training (frozen CLIP)}
\label{alg:procap}
\footnotesize
\setlength{\abovecaptionskip}{1pt}
\setlength{\belowcaptionskip}{0pt}
\begin{algorithmic}[1]
\State \textbf{Input:} $\mathcal{D}$; frozen $(E_I,E_T)$; depth $D$; $N_{\text{ctx}}$; weights $(\lambda,\beta,\gamma)$
\State \textbf{Input:} temps $(\tau,\tau_p)$; AdamW $(\eta,\omega)$; logvar clamp $[a,b]$
\State Freeze $(E_I,E_T)$; initialize $\Theta$ and $\tau$
\For{epoch $=1..E$}
  \For{$(x,y)\sim\mathcal{D}$}
    \State $\epsilon\sim\mathcal{N}(0,I)$
    \State $\log\sigma^2 \leftarrow \mathrm{clip}(\log\sigma^2,a,b)$;\;\; $\sigma \leftarrow \exp(0.5\log\sigma^2)$
    \State $c \leftarrow \mu + \sigma\odot\epsilon$ \Comment{inference: $c\!\leftarrow\!\mu$}
    \State Build prompts $T_0(y,c)$, $V_0(c)$
    \For{$\ell=1..D$}
      \State $\tilde{V}_{\ell-1}\leftarrow W_p V_{\ell-1}$
      \State $V_\ell \leftarrow V_{\ell-1} + \mathrm{MHA}_v(Q{=}V_{\ell-1},K{=}T_{\ell-1},V{=}T_{\ell-1})$
      \State $T_\ell \leftarrow T_{\ell-1} + \mathrm{MHA}_t(Q{=}T_{\ell-1},K{=}\tilde{V}_{\ell-1},V{=}\tilde{V}_{\ell-1})$
    \EndFor
    \State $v\leftarrow E_I(x;V_D)$;\;\; $t_k\leftarrow E_T(k;T_D)\;\forall k$
    \State $L_{\text{CE}}\leftarrow \mathrm{CE}(\cos(v,t_k)/\tau,\;y)$
    \State $z\leftarrow \mathrm{norm}(g_I(v))$;\;\; $c_k\leftarrow \mathrm{norm}(g_T(t_k))\;\forall k$
    \State $L_{\text{InfoNCE}}\leftarrow \mathrm{SymInfoNCE}(z,\{c_k\},y;\tau_p)$
    \State $L_{\text{KL}}\leftarrow \sum_j \mathrm{KL}(\mathcal{N}(\mu_j,\sigma_j^2)\,\|\,\mathcal{N}(0,1))$
    \State $L_2\leftarrow \sum_j \|\mu_j\|_2^2$
    \State $L\leftarrow L_{\text{CE}}+\lambda L_{\text{InfoNCE}}+\beta L_{\text{KL}}+\gamma L_2$
    \State Update $\Theta$ using AdamW$(\eta,\omega)$ \Comment{optional: AMP / grad clip}
  \EndFor
\EndFor
\State \textbf{Output:} trained ProCAP; inference uses $c=\mu$
\end{algorithmic}
\vspace{-1mm}
\end{algorithm}
We summarize the optimization of ProCAP in Alg.~\ref{alg:procap}. We update only the prompt parameters,
the cross-attention modules, and the lightweight projection heads, while keeping the CLIP backbone frozen.

\begingroup
\section{Experiments}
\label{sec:experiments}

\subsection{Experimental Setup}
\label{subsec:exp_setup}

\noindent We follow the standard three-setting evaluation protocol used in MaPLe-style prompt learning on
frozen CLIP backbones \cite{khattak2023maple,mistretta2024kdpl,alipour2024stylepro}. See Tables~\ref{tab:dataset_summary_supp} and~\ref{tab:templates_supp} for dataset statistics and prompt templates.

\medskip
\noindent\textbf{Base-to-Novel Generalization.}
For each dataset, classes are split into \emph{base} and \emph{novel} subsets. We train using only labeled
examples from the base classes (16-shot per class) and evaluate on both base and novel test classes
\cite{khattak2023maple,mistretta2024kdpl}. We report Top-1 accuracy on base (B) and novel (N) classes,
and summarize the base--novel trade-off using \HM{}:
\begin{equation}
\HM{} = \frac{2BN}{B+N}.
\label{eq:hm}
\end{equation}
We evaluate on 11 datasets: ImageNet, Caltech101, OxfordPets, StanfordCars, Flowers102, Food101,
FGVCAircraft, SUN397, UCF101, DTD, and EuroSAT \cite{khattak2023maple}.

\medskip
\noindent\textbf{Cross-Dataset Evaluation.}
We train prompts on ImageNet (16-shot) and directly evaluate on the remaining 10 target datasets without
any target fine-tuning. This setting measures how well the learned prompts transfer across datasets
\cite{khattak2023maple,alipour2024stylepro}.

\medskip
\noindent\textbf{Domain Generalization.}
We train on ImageNet and evaluate on four distribution-shifted ImageNet variants without target supervision:
ImageNetV2 \cite{recht2019imagenetv2}, ImageNet-Sketch \cite{wang2019learning}, ImageNet-A \cite{hendrycks2019natural},
and ImageNet-R \cite{hendrycks2021many}. This setting measures robustness under real distribution shift
\cite{khattak2023maple}.

\medskip
\noindent\textbf{Implementation Details.}
We use CLIP ViT-B/16 \cite{radford2021clip} and keep both image and text encoders frozen. We optimize only
prompt parameters, ProCAP cross-attention modules, and lightweight projection heads used by auxiliary losses.
All images are resized to $224\times224$ with bicubic interpolation and normalized using CLIP’s mean/std
\cite{radford2021clip}. Training uses random resized crop (scale 0.8--1.0) and horizontal flip; for some
datasets (e.g., EuroSAT) we additionally apply mild color jitter and small random rotations. These are training-time augmentations only; no horizontal-flip test-time augmentation is used.
We use AdamW with cosine learning-rate schedule and linear warmup. Unless stated otherwise, we train for
50 epochs with batch size 32 (train) / 100 (test), learning rate $2.5\times10^{-4}$, weight decay 0.03,
warmup 3 epochs, and AMP mixed precision. For the main base-to-novel results, we report averages over three random seeds. For auxiliary ablations and runtime measurements, we use seed 1 unless otherwise stated.
For ProCAP, we use $N_{\text{ctx}}{=}4$ learnable context tokens (motivated by the prompt-length ablation in
Figure~\ref{fig:prompt_length}) and initialize prompts with dataset-specific templates (e.g., EuroSAT:
``a centered satellite photo of a''); multiple templates are specified in the configuration using the delimiter ``||'', which is used only by the parser to separate templates before tokenization.
Prompt injection depth is nine prompted transformer layers in the main configuration. Cross-modal interaction
is implemented using the bidirectional residual multi-head cross-attention described in Section~\ref{subsec:procapprompts}.
The full ProCAP configuration uses: (i) a symmetric InfoNCE head (projection dim 256, temperature init $\tau_{\mathrm{p}}=0.07$, weight $\lambda=0.01$),
(ii) Gaussian prompt-space regularization ($\beta=10^{-5}$ with log-variance clamping),
and (iii) an extra prompt-only $L_2$ penalty (weight $10^{-6}$). See Tables~\ref{tab:eff_runtime_supp} and~\ref{tab:procap_epochs_supp} for runtime and training-duration analyses.

\medskip
\noindent\textbf{Run-to-run stability.}
To assess robustness to random initialization, we repeat ProCAP training with three random seeds under the 16-shot base-to-novel setting. Full mean $\pm$ standard deviation results for base, novel, and \HM{} across all 11 datasets are reported in Table~\ref{tab:procap_seed_stability_supp}. On average, ProCAP achieves 85.43$\pm$0.18 base and 77.61$\pm$0.31 novel accuracy. The arithmetic mean of the 11 dataset-wise HMs is 81.12$\pm$0.24; the aggregate HM computed from the averaged base and novel accuracies, as in Table~\ref{tab:base_to_novel}, is 81.33. These results indicate stable convergence across runs. The largest variations occur on more sensitive datasets such as DTD, EuroSAT, and FGVCAircraft.

\subsection{Main Results}
\label{subsec:main_results}

\noindent\textbf{Comparison methods and sources.}
We use results reported in the original publications for CLIP~\citep{radford2021clip}, CoOp and CoCoOp~\citep{zhou2022coop,zhou2022cocoop}, KgCoOp~\citep{yao2023kgcoop}, MaPLe~\citep{khattak2023maple}, PromptSRC~\citep{khattak2023selfreg}, CoPrompt~\citep{roy2024coprompt}, TCP~\citep{yang2024tcp}, MMA~\citep{chen2024mma}, 2SFS~\citep{farina2025two}, SkipT~\citep{zhu2025skiptuning}, SPTR~\citep{sptr2025aaai}, Sparse-KgCoOp~\citep{sparsekgcoop2025}, MMRL~\citep{guo2025mmrl}, and the recent HI$^{2}$MA, DCPL, MPCLe, and MMLoP methods~\citep{dong2026hi2ma,mei2026dcpl,zhang2026mpcle,ghiasvand2026mmlop}. Unreported values are left as dashes.

\noindent\textbf{Base-to-Novel Results.}
Table~\ref{tab:base_to_novel} compares ProCAP with established methods and the recent methods HI$^{2}$MA, DCPL, MPCLe, and MMLoP. ProCAP obtains an aggregate average \HM{} of 81.33. Relative to MaPLe, ProCAP improves \HM{} on
nearly all benchmarks, with especially clear gains on shift- and texture-sensitive datasets such as
EuroSAT (90.98 vs.\ 82.35 HM) and DTD (73.17 vs.\ 68.16 HM). We also observe consistent improvements on
fine-grained recognition (e.g., OxfordPets: 97.34 vs.\ 96.58 HM) and several general benchmarks
(e.g., Caltech101: 97.11 vs.\ 96.02 HM), while keeping base accuracy competitive. These results suggest
that tighter cross-modal coupling and low-shot regularization help ProCAP generalize more reliably to
novel classes. For the 2026 entries, we transcribe only the metrics reported by the corresponding cited sources; unreported cells are left as dashes.

\noindent\textbf{Domain Generalization.}
Table~\ref{tab:domain_gen} reports performance on ImageNet distribution shifts and includes the available
results for the recent 2026 methods. ProCAP obtains 72.14 on the source ImageNet split and a target mean of
60.58 across ImageNetV2, ImageNet-Sketch, ImageNet-A, and ImageNet-R. Only metrics explicitly reported by
each cited source are transcribed; a dash denotes an unreported cell.

\noindent\textbf{Cross-Dataset Results.}
Table~\ref{tab:cross_dataset} summarizes cross-dataset transfer performance and integrates the available 2026 results from the cited sources. ProCAP records an average accuracy of 70.02 across the 10 target datasets, with strong results on DTD, EuroSAT, SUN397, and UCF101, indicating robust transfer under dataset shift.

\begin{table}[H]
\centering
\caption{Base-to-novel generalization (Top-1 \%, HM). In the Average block, HM is computed from the average Base and Novel accuracies using Eq.~\eqref{eq:hm}. Dashes indicate unreported metrics; ProCAP reports three-seed means.}
\label{tab:base_to_novel}
\begingroup
\setlength{\tabcolsep}{6.5pt}
\renewcommand{\arraystretch}{0.77}
\scriptsize

\resizebox{\textwidth}{!}{%
\begin{tabular}{llccc|ccc|ccc|ccc}
\toprule
\multirow{2}{*}{Method} & \multirow{2}{*}{Venue}
& \multicolumn{3}{c|}{(a) Average}
& \multicolumn{3}{c|}{(b) ImageNet}
& \multicolumn{3}{c|}{(c) Caltech101}
& \multicolumn{3}{c}{(d) OxfordPets} \\
\cmidrule(lr){3-5}\cmidrule(lr){6-8}\cmidrule(lr){9-11}\cmidrule(lr){12-14}
& & Base & Novel & HM
& Base & Novel & HM
& Base & Novel & HM
& Base & Novel & HM \\
\midrule
CLIP  & ICML'21
& 69.34 & 74.22 & 71.70 & 72.43 & 68.14 & 70.22 & 96.84 & 94.00 & 95.40 & 91.17 & 97.26 & 94.12 \\
CoCoOp  & CVPR'22
& 80.47 & 71.69 & 75.83 & 75.98 & 70.43 & 73.10 & 97.96 & 93.81 & 95.84 & 95.20 & 97.69 & 96.43 \\
CoOp  & IJCV'22
& 82.69 & 63.22 & 71.66 & 76.47 & 67.88 & 71.92 & 98.00 & 89.81 & 93.73 & 93.67 & 95.29 & 94.47 \\
KgCoOp  & CVPR'23
& 80.73 & 73.60 & 77.00 & 75.83 & 69.96 & 72.78 & 97.72 & 94.39 & 96.03 & 94.65 & 97.76 & 96.18 \\
MaPLe  & CVPR'23
& 82.28 & 75.14 & 78.55 & 76.66 & 70.54 & 73.47 & 97.74 & 94.36 & 96.02 & 95.43 & 97.76 & 96.58 \\
PromptSRC  & ICCV'23
& 84.26 & 76.10 & 79.97 & 77.60 & 70.73 & 74.01 & 98.10 & 94.03 & 96.02 & 95.33 & 97.30 & 96.30 \\
CoPrompt  & ICLR'24
& 84.00 & 77.23 & 80.47 & 77.67 & 71.27 & 74.33 & 98.27 & 94.90 & 96.56 & 95.67 & 98.10 & 96.87 \\
TCP  & CVPR'24
& 84.13 & 75.36 & 79.50 & 77.27 & 69.87 & 73.38 & 98.23 & 94.67 & 96.42 & 94.67 & 97.20 & 95.92 \\
MMA  & CVPR'24
& 83.20 & 76.80 & 79.87 & 77.31 & 71.00 & 74.02 & 98.40 & 94.00 & 96.15 & 95.40 & 98.07 & 96.72 \\
2SFS  & CVPR'25
& \best{85.55} & 75.48 & 80.20 & 77.71 & 70.99 & 74.20 & \second{98.71} & 94.43 & 96.52 & 95.32 & 97.82 & 96.55 \\
SkipT  & CVPR'25
& 85.04 & 77.53 & \second{81.11} & \second{77.73} & 70.40 & 73.89 & 98.50 & \best{95.33} & \second{96.89} & 95.70 & 97.87 & 96.77 \\
SPTR  & AAAI'25
& 84.85 & 76.76 & 80.60 & 77.60 & \best{71.75} & \best{74.56} & 98.01 & 93.83 & 95.87 & 94.85 & 97.10 & 95.97 \\
Sparse-KgCoOp  & Entropy'25
& 83.20 & 73.71 & 78.17 & 76.68 & 69.01 & 72.64 & 97.33 & 93.43 & 95.34 & 93.89 & \best{100.00} & 96.90 \\
HI$^{2}$MA  & PR'26
& 84.24 & \best{77.96} & 80.98
& \best{78.10} & \second{71.30} & \second{74.55}
& 98.53 & 95.20 & 96.84
& \second{96.13} & 97.83 & \second{96.97} \\
DCPL  & CAAI TIT'26
& 85.03 & 77.01 & 80.82 & -- & -- & -- & -- & -- & -- & -- & -- & -- \\
MPCLe  & Inf. Sci.'26
& 84.91 & 76.61 & 80.55
& 77.57 & 70.83 & 74.05
& 98.33 & 95.17 & 96.72
& 96.03 & 97.77 & 96.89 \\
MMLoP  & ECCV'26
& 83.79 & 75.98 & 79.69 & 77.00 & 70.50 & 73.61 & 98.27 & 93.93 & 96.05 & 95.47 & 96.97 & 96.21 \\
\rowcolor{gray!15}
ProCAP & Ours
& \second{85.43} & \second{77.61} & \best{81.33}
& 77.33 & 70.00 & 73.48
& \best{98.97} & \second{95.31} & \best{97.11}
& \best{96.49} & \second{98.21} & \best{97.34} \\
\bottomrule
\end{tabular}}

\vspace{0pt}

\resizebox{\textwidth}{!}{%
\begin{tabular}{llccc|ccc|ccc|ccc}
\toprule
\multirow{2}{*}{Method} & \multirow{2}{*}{Venue}
& \multicolumn{3}{c|}{(e) StanfordCars}
& \multicolumn{3}{c|}{(f) Flowers102}
& \multicolumn{3}{c|}{(g) Food101}
& \multicolumn{3}{c}{(h) FGVCAircraft} \\
\cmidrule(lr){3-5}\cmidrule(lr){6-8}\cmidrule(lr){9-11}\cmidrule(lr){12-14}
& & Base & Novel & HM
& Base & Novel & HM
& Base & Novel & HM
& Base & Novel & HM \\
\midrule
CLIP  & ICML'21
& 63.37 & 74.89 & 68.65 & 72.08 & \best{77.80} & 74.83 & 90.10 & 91.22 & 90.66 & 27.19 & 36.29 & 31.09 \\
CoCoOp  & CVPR'22
& 70.49 & 73.59 & 72.01 & 94.87 & 71.75 & 81.71 & 90.70 & 91.29 & 90.99 & 33.41 & 23.71 & 27.74 \\
CoOp  & IJCV'22
& 78.12 & 60.40 & 68.13 & 97.60 & 59.67 & 74.06 & 88.33 & 82.26 & 85.19 & 40.44 & 22.30 & 28.75 \\
KgCoOp  & CVPR'23
& 71.76 & 75.04 & 73.36 & 95.00 & 74.73 & 83.65 & 90.50 & 91.70 & 91.09 & 36.21 & 33.55 & 34.83 \\
MaPLe  & CVPR'23
& 72.94 & 74.00 & 73.47 & 95.92 & 72.46 & 82.56 & 90.71 & 92.05 & 91.38 & 37.44 & 35.61 & 36.50 \\
PromptSRC  & ICCV'23
& 78.27 & 74.97 & 76.58 & 98.07 & 76.50 & 85.95 & 90.67 & 91.53 & 91.10 & 42.73 & 37.87 & 40.15 \\
CoPrompt  & ICLR'24
& 76.97 & 74.40 & 75.66 & 97.27 & 76.60 & 85.71 & 90.73 & 92.07 & 91.40 & 40.20 & \second{39.33} & 39.76 \\
TCP  & CVPR'24
& 80.80 & 74.13 & 77.32 & 97.73 & 75.57 & 85.23 & 90.57 & 91.37 & 90.97 & 41.97 & 34.43 & 37.83 \\
MMA  & CVPR'24
& 78.50 & 73.10 & 75.70 & 97.77 & 75.93 & 85.48 & 90.13 & 91.30 & 90.71 & 40.57 & 36.33 & 38.33 \\
2SFS  & CVPR'25
& \second{82.50} & 74.80 & \second{78.46} & 98.29 & 76.17 & 85.83 & 89.11 & 91.34 & 90.21 & \best{47.48} & 35.51 & 40.63 \\
SkipT  & CVPR'25
& \best{82.93} & 72.50 & 77.37 & \best{98.57} & 75.80 & 85.70 & 90.67 & 92.03 & 91.34 & 45.37 & 37.13 & 40.84 \\
SPTR  & AAAI'25
& 81.85 & \second{75.43} & \best{78.53} & \second{98.56} & \second{77.59} & \best{86.86} & \best{91.11} & \best{92.74} & \best{91.94} & 44.26 & \best{40.18} & \best{42.09} \\
Sparse-KgCoOp  & Entropy'25
& 78.86 & 74.09 & 76.39 & 96.14 & 74.51 & 83.95 & 89.78 & \second{92.50} & 91.12 & 42.37 & 33.26 & 37.24 \\
HI$^{2}$MA  & PR'26
& 80.67 & 74.47 & 77.45
& 97.83 & 77.07 & 86.22
& \second{90.97} & 92.07 & \second{91.52}
& 43.20 & 38.97 & 40.98 \\
MPCLe  & Inf. Sci.'26
& 81.17 & 74.90 & 77.91
& 98.13 & 77.03 & \second{86.31}
& 90.80 & 92.00 & 91.40
& 43.37 & 37.03 & \second{41.57} \\
MMLoP  & ECCV'26
& 77.77 & 74.83 & 76.27 & 97.63 & 76.73 & 85.93 & 90.70 & 91.70 & 91.19 & 42.17 & 34.60 & 38.01 \\
\rowcolor{gray!15}
ProCAP & Ours
& 78.65 & \best{77.26} & 77.95
& 98.39 & 75.60 & 85.50
& 90.60 & 92.10 & 91.34
& \second{45.92} & 36.77 & 40.84 \\
\bottomrule
\end{tabular}}

\vspace{0pt}

\resizebox{\textwidth}{!}{%
\begin{tabular}{llccc|ccc|ccc|ccc}
\toprule
\multirow{2}{*}{Method} & \multirow{2}{*}{Venue}
& \multicolumn{3}{c|}{(i) SUN397}
& \multicolumn{3}{c|}{(j) DTD}
& \multicolumn{3}{c|}{(k) EuroSAT}
& \multicolumn{3}{c}{(l) UCF101} \\
\cmidrule(lr){3-5}\cmidrule(lr){6-8}\cmidrule(lr){9-11}\cmidrule(lr){12-14}
& & Base & Novel & HM
& Base & Novel & HM
& Base & Novel & HM
& Base & Novel & HM \\
\midrule
CLIP  & ICML'21
& 69.36 & 75.35 & 72.23 & 53.24 & 59.90 & 56.37 & 56.48 & 64.05 & 60.03 & 70.53 & 77.50 & 73.85 \\
CoCoOp  & CVPR'22
& 79.74 & 76.86 & 78.27 & 77.01 & 56.00 & 64.85 & 87.49 & 60.04 & 71.21 & 82.33 & 73.45 & 77.64 \\
CoOp  & IJCV'22
& 80.60 & 65.89 & 72.51 & 79.44 & 41.18 & 54.24 & 92.19 & 54.74 & 68.69 & 84.69 & 56.05 & 67.46 \\
KgCoOp  & CVPR'23
& 80.29 & 76.53 & 78.36 & 77.55 & 54.99 & 64.35 & 85.64 & 64.34 & 73.48 & 82.89 & 76.67 & 79.65 \\
MaPLe  & CVPR'23
& 80.82 & 78.70 & 79.75 & 80.36 & 59.18 & 68.16 & 94.07 & 73.23 & 82.35 & 83.00 & 78.66 & 80.77 \\
PromptSRC  & ICCV'23
& 82.67 & 78.47 & 80.52 & 83.37 & 62.97 & 71.75 & 92.90 & 73.90 & 82.32 & 87.10 & 78.80 & 82.74 \\
CoPrompt  & ICLR'24
& 82.63 & \second{80.03} & \second{81.31} & 83.13 & 64.73 & 72.79 & 94.60 & 78.57 & 85.84 & 86.90 & 79.57 & 83.07 \\
TCP  & CVPR'24
& 82.63 & 78.20 & 80.35 & 82.77 & 58.07 & 68.25 & 91.63 & 74.73 & 82.32 & 87.13 & 80.77 & 83.83 \\
MMA  & CVPR'24
& 82.27 & 78.57 & 80.38 & 83.20 & 65.63 & 73.38 & 85.46 & 82.34 & 83.87 & 86.23 & 80.03 & 83.01 \\
2SFS  & CVPR'25
& 82.59 & 78.91 & 80.70 & \second{84.60} & 65.01 & 73.52 & \second{96.91} & 67.09 & 79.29 & 87.85 & 78.19 & 82.74 \\
SkipT  & CVPR'25
& 82.40 & 79.03 & 80.68 & 83.77 & \best{67.23} & \best{74.59} & 92.47 & \second{83.00} & \second{87.48} & 87.30 & \best{82.47} & \best{84.81} \\
SPTR  & AAAI'25
& 82.56 & 79.26 & 80.90 & 83.35 & 62.58 & 71.50 & 93.15 & 72.80 & 81.73 & \best{88.17} & 81.00 & 84.47 \\
Sparse-KgCoOp  & Entropy'25
& 81.60 & 74.92 & 78.12 & 81.80 & 54.94 & 65.73 & 93.61 & 68.61 & 79.18 & 84.73 & 75.39 & 79.79 \\
HI$^{2}$MA  & PR'26
& \best{83.30} & \best{80.13} & \best{81.68}
& 84.33 & \second{66.30} & \second{74.24}
& 86.63 & 81.83 & 84.16
& 86.97 & \second{82.43} & \second{84.64} \\
MPCLe  & Inf. Sci.'26
& \second{83.17} & 78.90 & 80.98
& 82.77 & 60.37 & 69.82
& 95.37 & 78.93 & 86.37
& 87.40 & 79.77 & 83.41 \\
MMLoP  & ECCV'26
& 82.40 & 78.13 & 80.21 & 82.67 & 60.87 & 70.11 & 92.37 & 79.10 & 85.22 & 85.23 & 78.37 & 81.66 \\
\rowcolor{gray!15}
ProCAP & Ours
& 81.90 & 79.45 & 80.66
& \best{85.19} & 64.13 & 73.17
& \best{98.21} & \best{84.74} & \best{90.98}
& \second{88.11} & 80.10 & 83.91 \\
\bottomrule
\end{tabular}}
\endgroup
\end{table}

\begin{table}[!t]
\caption{Domain generalization (Top-1 \%). Train on ImageNet and evaluate directly on ImageNet distribution shifts. Dashes indicate unreported values. DCPL$^{*}$ denotes the domain-generalization variant defined by \citet{mei2026dcpl}: its image-specific prompt is used only as an auxiliary input, while CLIP's original image classification token is retained.}
\label{tab:domain_gen}
\small
\centering
\setlength{\tabcolsep}{5pt}
\renewcommand{\arraystretch}{1.15}
\resizebox{\textwidth}{!}{%
\begin{tabular}{llcccccc}
\toprule
\multirow{2}{*}{Method} & \multirow{2}{*}{Venue} & \multicolumn{1}{c}{Source} & \multicolumn{4}{c}{Target} & \multicolumn{1}{c}{Target Mean} \\
\cmidrule(lr){3-3}\cmidrule(lr){4-7}\cmidrule(lr){8-8}
& & ImageNet & ImageNetV2 & ImageNet-Sketch & ImageNet-A & ImageNet-R & Avg. \\
\midrule
CLIP  & ICML'21 & 66.73 & 60.83 & 46.15 & 47.77 & 73.96 & 57.18 \\
CoOp  & IJCV'22 & 71.51 & 64.20 & 47.99 & 49.71 & 75.21 & 59.28 \\
CoCoOp  & CVPR'22 & 71.02 & 64.07 & 48.75 & 50.63 & 76.18 & 59.91 \\
MaPLe  & CVPR'23 & 70.72 & 64.07 & 49.15 & 50.90 & 76.98 & 60.28 \\
PromptSRC  & ICCV'23 & 71.27 & 64.35 & 49.55 & 50.90 & \best{77.80} & 60.65 \\
MMRL  & CVPR'25 & \second{72.03} & 64.47 & 49.17 & 51.20 & 77.53 & 60.59 \\
HI$^{2}$MA & PR'26 & 71.87 & \best{65.97} & \best{49.73} & 50.13 & 77.30 & \second{60.78} \\
DCPL$^{*}$ & CAAI TIT'26 & 71.37 & \second{64.73} & 49.50 & \second{51.37} & 77.53 & \second{60.78} \\
MPCLe & Inf. Sci.'26 & 71.33 & 64.57 & \second{49.67} & \best{51.57} & \second{77.63} & \best{60.86} \\
MMLoP & ECCV'26 & 71.00 & 64.30 & 49.07 & 50.83 & \second{77.63} & 60.46 \\
\rowcolor{gray!15}
ProCAP    & Ours    & \best{72.14} & 64.60 & 49.53 & 50.83 & 77.35 & 60.58 \\
\bottomrule
\end{tabular}}
\end{table}


\begin{table}[!t]
\caption{Cross-dataset transfer (Top-1 \%). Train on ImageNet (16-shot) and evaluate directly on 10 target datasets.}
\label{tab:cross_dataset}
\scriptsize

\centering
\setlength{\tabcolsep}{2.9pt}
\renewcommand{\arraystretch}{1.12}
\resizebox{\textwidth}{!}{%
\begin{tabular}{lcccccccccccc}
\toprule
\multirow{2}{*}{Method} & \multicolumn{1}{c}{Source} & \multicolumn{11}{c}{Target} \\
\cmidrule(lr){2-2}\cmidrule(lr){3-13}
& ImageNet & Avg. & Caltech101 & Pets & Cars & Flowers & Food & Aircraft & SUN397 & DTD & EuroSAT & UCF101 \\
\midrule
CoOp  & 71.51 & 63.88 & 93.70 & 89.14 & 64.51 & 68.71 & 85.30 & 18.47 & 64.15 & 41.92 & 46.39 & 66.55 \\
CoCoOp  & 71.02 & 65.74 & 94.43 & 90.14 & 65.32 & 71.88 & 86.06 & 22.94 & 67.36 & 45.73 & 45.37 & 68.21 \\
MaPLe  & 70.72 & 66.30 & 93.53 & 90.49 & 65.57 & 72.23 & 86.20 & 24.74 & 67.01 & 46.49 & 48.06 & 68.69 \\
PromptSRC  & 71.27 & 65.81 & 93.60 & 90.25 & 65.70 & 70.25 & 86.15 & 23.90 & 67.10 & 46.87 & 45.50 & 68.75 \\
TCP  & 71.40 & 66.29 & 93.97 & \second{91.25} & 64.69 & 71.21 & \second{86.69} & 23.45 & 67.15 & 44.35 & 51.45 & 68.73 \\
MMRL  & \second{72.03} & \second{67.25} & 94.67 & \best{91.43} & \second{66.10} & \second{72.77} & 86.40 & \second{26.30} & 67.57 & 45.90 & \second{53.10} & 68.27 \\
HI$^{2}$MA  & 71.87 & 66.83 & 94.53 & 90.30 & \best{66.17} & 72.23 & 86.23 & 25.23 & \second{67.73} & 47.73 & 48.40 & \second{69.77} \\
DCPL  & 70.63 & 66.53 & 94.03 & 90.23 & 65.57 & 70.07 & 86.17 & 24.00 & 67.17 & 47.27 & \second{53.10} & 67.70 \\
MPCLe  & 71.33 & 67.02 & \second{94.77} & 90.63 & 65.83 & 72.43 & 86.47 & 25.27 & 67.43 & \second{47.83} & 49.87 & 69.67 \\
\rowcolor{gray!15}
ProCAP (Ours) & \best{72.14} & \best{70.02} & \best{96.60} & 91.20 & 65.10 & \best{72.90} & \best{89.10} & \best{27.70} & \best{71.10} & \best{54.20} & \best{57.70} & \best{74.60} \\
\bottomrule
\end{tabular}}
\end{table}

\subsection{Few-shot Evaluation}
\label{subsec:fewshot_eval}

\noindent
We evaluate $K$-shot classification with $K\in\{1,2,4,8,16\}$ labeled examples per class and report
Top-1 accuracy (\%) on the test split across datasets. Figure~\ref{fig:shots} summarizes how
performance scales with increasing supervision.

\begin{figure}[!t]
  \centering
\includegraphics[width=\textwidth,trim=0 12 0 10,clip]{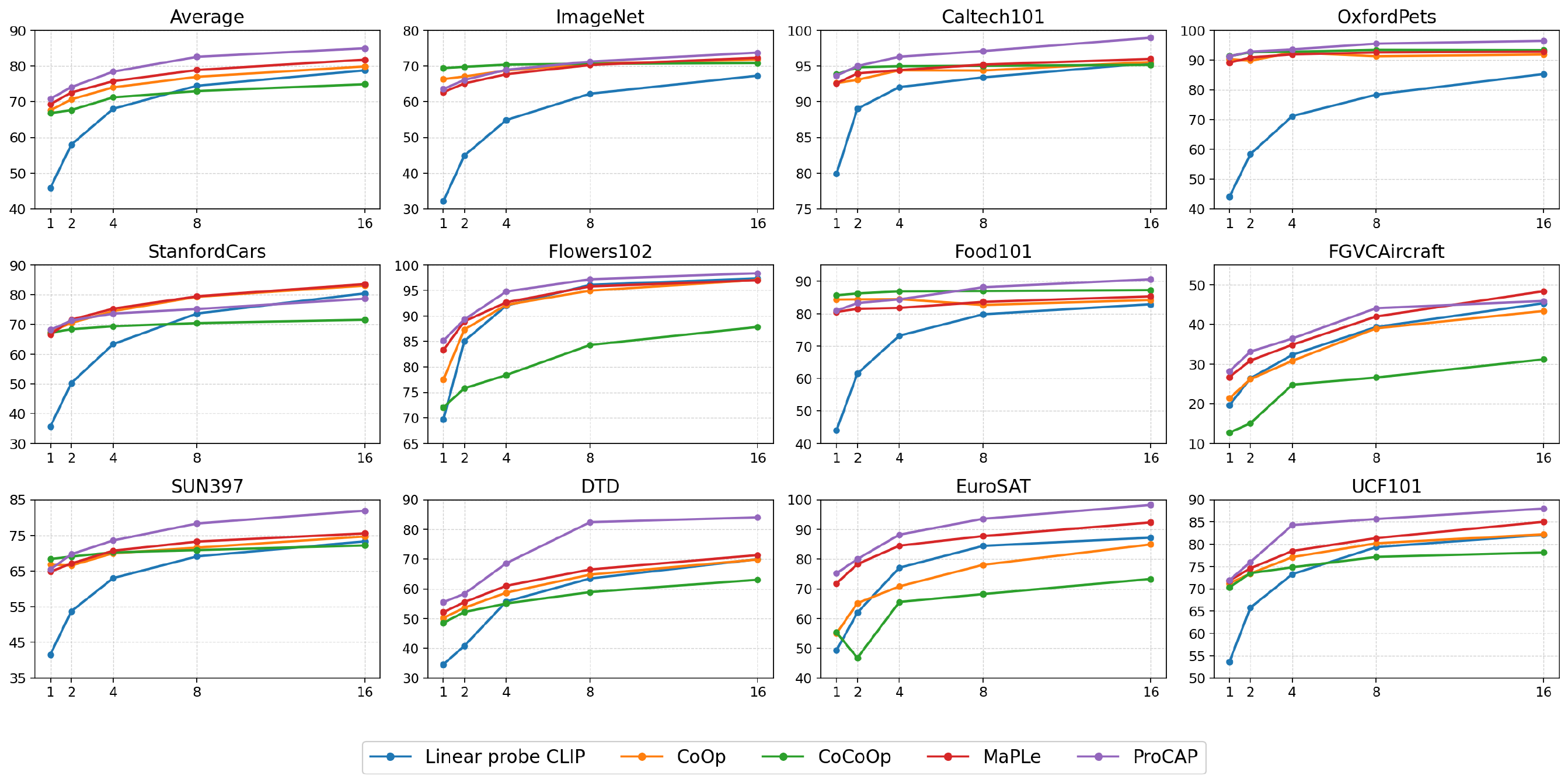}
  \captionsetup{skip=3pt}
  \caption{Few-shot learning curves (Top-1 accuracy, \%) across benchmark datasets with
  $K\in\{1,2,4,8,16\}$ training shots per class (CLIP ViT-B/16). We compare ProCAP against standard
  prompt-learning baselines under the same protocol. The \emph{Average} panel reports the mean
  accuracy over all evaluated datasets.}
  \label{fig:shots}
\end{figure}

\noindent
As expected, accuracy improves monotonically as more labeled examples become available.
Across datasets, ProCAP exhibits strong data efficiency, with the clearest advantages in the
low-shot regime (1--4 shots), where effective regularization and cross-modal coupling matter most.
As $K$ increases, the gap to competitive baselines typically narrows, but ProCAP remains consistently
among the top performers, indicating stable scaling behavior rather than gains limited to a single
shot setting. Notably, the relative improvements are more pronounced on challenging fine-grained and
domain-shifted benchmarks, suggesting that the learned prompts transfer robustly when supervision is scarce.

\subsection{Ablation Study}\label{sec:ablation}

\newcommand{\cmark}{\checkmark}
\newcommand{\xmark}{--}

\noindent We analyze the contribution of each ProCAP component via controlled ablations.

\begin{table}[H]
\centering
\caption{Ablation study of ProCAP components (CLIP ViT-B/16, 16-shot). We report Top-1 accuracy (\%) on representative texture, remote-sensing, and fine-grained datasets.}
\label{tab:ablation_components}
\scriptsize
\setlength{\tabcolsep}{3.2pt}
\renewcommand{\arraystretch}{1.12}
\resizebox{\textwidth}{!}{%
\begin{tabular}{p{2.8cm}ccccc|cccc|c}
\toprule
Variant & Cross-Attn & Gaussian & KL & $L_2$ & InfoNCE
& DTD & EuroSAT & OxfordPets & StanfordCars & Avg. \\
\midrule
MaPLe-style \tabcite{khattak2023maple}
& \xmark & \xmark & \xmark & \xmark & \xmark
& 67.65 & 80.15 & 96.58 & 73.47 & 79.46 \\

w/o Cross-Attn
& \xmark & \cmark & \cmark & \cmark & \cmark
& 72.86 & 90.22 & 96.67 & 77.10 & 84.21 \\

w/o Gaussian
& \cmark & \xmark & \xmark & \xmark & \cmark
& 72.35 & 89.92 & 96.60 & 76.80 & 83.92 \\

w/o KL
& \cmark & \cmark & \xmark & \cmark & \cmark
& 72.88 & 90.12 & 97.05 & 77.55 & 84.40 \\

w/o $L_2$
& \cmark & \cmark & \cmark & \xmark & \cmark
& \second{73.05} & \second{90.55} & 96.44 & \second{77.90} & \second{84.49} \\

w/o InfoNCE
& \cmark & \cmark & \cmark & \cmark & \xmark
& 72.32 & 89.46 & \second{97.13} & 77.30 & 84.05 \\

\midrule
\rowcolor{gray!12}
Full ProCAP
& \cmark & \cmark & \cmark & \cmark & \cmark
& \best{73.17} & \best{90.98} & \best{97.34} & \best{77.95} & \best{84.86} \\
\bottomrule
\end{tabular}}
\end{table}

\noindent
Table~\ref{tab:ablation_components} shows that each ProCAP component contributes positively across representative datasets. In addition to DTD and EuroSAT, we include OxfordPets and StanfordCars to verify that the ablation trend is not limited to texture or remote-sensing benchmarks. Removing any single component reduces the average performance. The largest average drop is observed when removing Gaussian prompt modeling, followed by removing the InfoNCE head and cross-attention, indicating that prompt-space regularization, image--class alignment, and bidirectional visual--text interaction all contribute to the final result.

\begin{figure}[!tbh]
  \centering
\includegraphics[width=0.75\columnwidth]{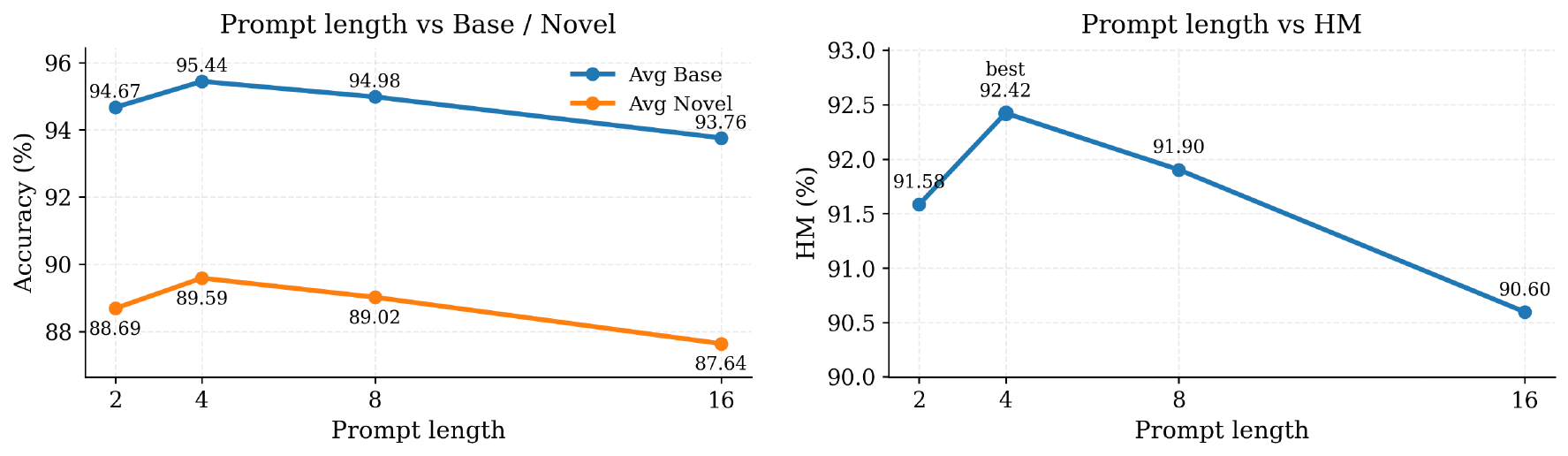}
  \caption{Effect of prompt length ($N_{\text{ctx}}$) on base-to-novel generalization (averaged over EuroSAT, Caltech101, OxfordPets, and UCF101).
  Left: average base and novel accuracies. Right: harmonic mean (HM). Performance peaks at $N_{\text{ctx}}{=}4$, indicating that a compact context is sufficient, while longer prompts tend to overfit under low-shot supervision.}
  \label{fig:prompt_length}
\end{figure}

\noindent
We study the effect of prompt length $N_{\text{ctx}}$. As shown in Fig.~\ref{fig:prompt_length}, performance peaks at
$N_{\text{ctx}}{=}4$: increasing from 2 to 4 improves base/novel accuracy and HM, while longer prompts (8, 16) tend to
overfit and reduce HM. We therefore use $N_{\text{ctx}}{=}4$ in all experiments.

\noindent
We further analyze robustness to key hyperparameters.
Figure~\ref{fig:sensitivity} reports \HM{} sensitivity to the contrastive loss weight $\lambda$,
uncertainty regularization weight $\beta$, and InfoNCE temperature initialization $\tau_{\mathrm{p}}$, showing stable performance across a broad range.
The selected configuration is $\lambda{=}0.01$, $\beta{=}10^{-5}$, and $\tau_{\mathrm{p}}{=}0.07$.

\begin{figure}[!t]
  \centering
\includegraphics[width=0.8\columnwidth]{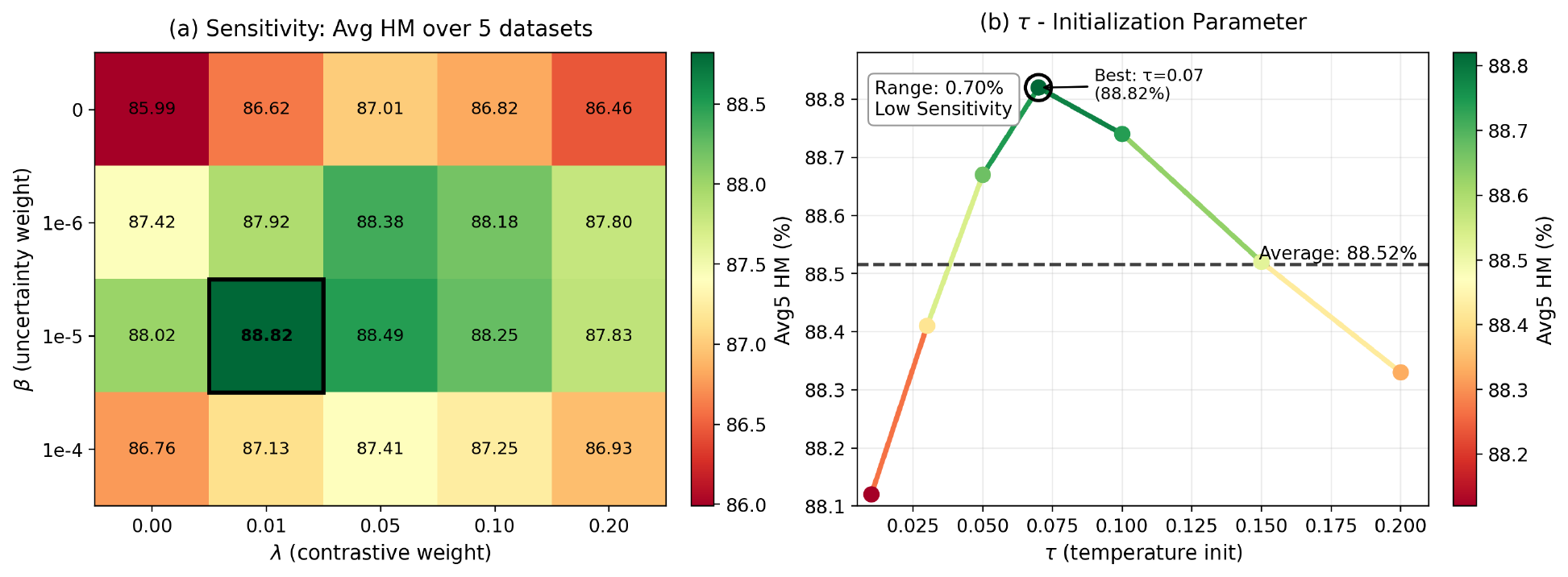}
  \caption{Hyperparameter sensitivity of ProCAP. \HM{} averaged over EuroSAT, OxfordPets, Caltech101, DTD, and Flowers102.
  \textbf{(a)} Sensitivity to $\lambda$ and $\beta$. \textbf{(b)} Sensitivity to InfoNCE temperature init $\tau_{\mathrm{p}}$ (shown as $\tau$ in the plotted sweep) with $\lambda,\beta$ fixed.}
  \label{fig:sensitivity}
\end{figure}

Additional prompt-token visualizations are provided in Appendix~\ref{sec:additional_results}.

\endgroup

\section{Conclusion}
\label{sec:conclusion}

We presented \textbf{ProCAP}, a probabilistic cross-attentive multimodal prompt learner for adapting frozen CLIP models under limited supervision and distribution shift. ProCAP learns visual and textual prompt tokens and couples them via stacked bidirectional multi-head cross-attention across prompt depth, enabling stronger two-way refinement than MaPLe-style prompting. To stabilize low-shot optimization, we parameterize prompts as Gaussians with lightweight KL/$L_2$ regularization and add a compact symmetric InfoNCE head to better align cross-attended image features with class-level text representations. Experiments on 11 datasets, cross-dataset transfer, and ImageNet shift benchmarks show strong aggregate base-to-novel performance and competitive cross-dataset and domain-transfer performance while keeping CLIP fully frozen. Future work will explore more efficient cross-attention designs and scaling to larger vision--language backbones.

\section*{Broader Impact}
ProCAP is a general-purpose prompt-learning method for image classification and does not introduce new data collection or human-subject experiments. As with other CLIP-based systems, its predictions may inherit biases, dataset imbalances, and representation limitations from the pretrained model and evaluation benchmarks. Deployment in high-stakes settings may therefore amplify errors or unequal performance across populations. Appropriate mitigations include application-specific dataset auditing, subgroup evaluation, uncertainty calibration, and human oversight; the method should not be used as the sole basis for safety-critical decisions.
\begingroup
\setlength{\bibsep}{4pt}
\bibliography{main}
\bibliographystyle{tmlr}
\endgroup

\raggedbottom
\appendix
\input{additional_material}

\end{document}

%% file: additional_material.tex
\section{Additional Experimental Results}
\label{sec:additional_results}

This appendix provides qualitative comparisons, three-seed stability analysis, runtime measurements, reproducibility details, dataset statistics, and the prompt templates used in our experiments.

\subsection{Qualitative Comparison}

\begin{figure}[H]
  \centering
  \includegraphics[width=0.72\textwidth]{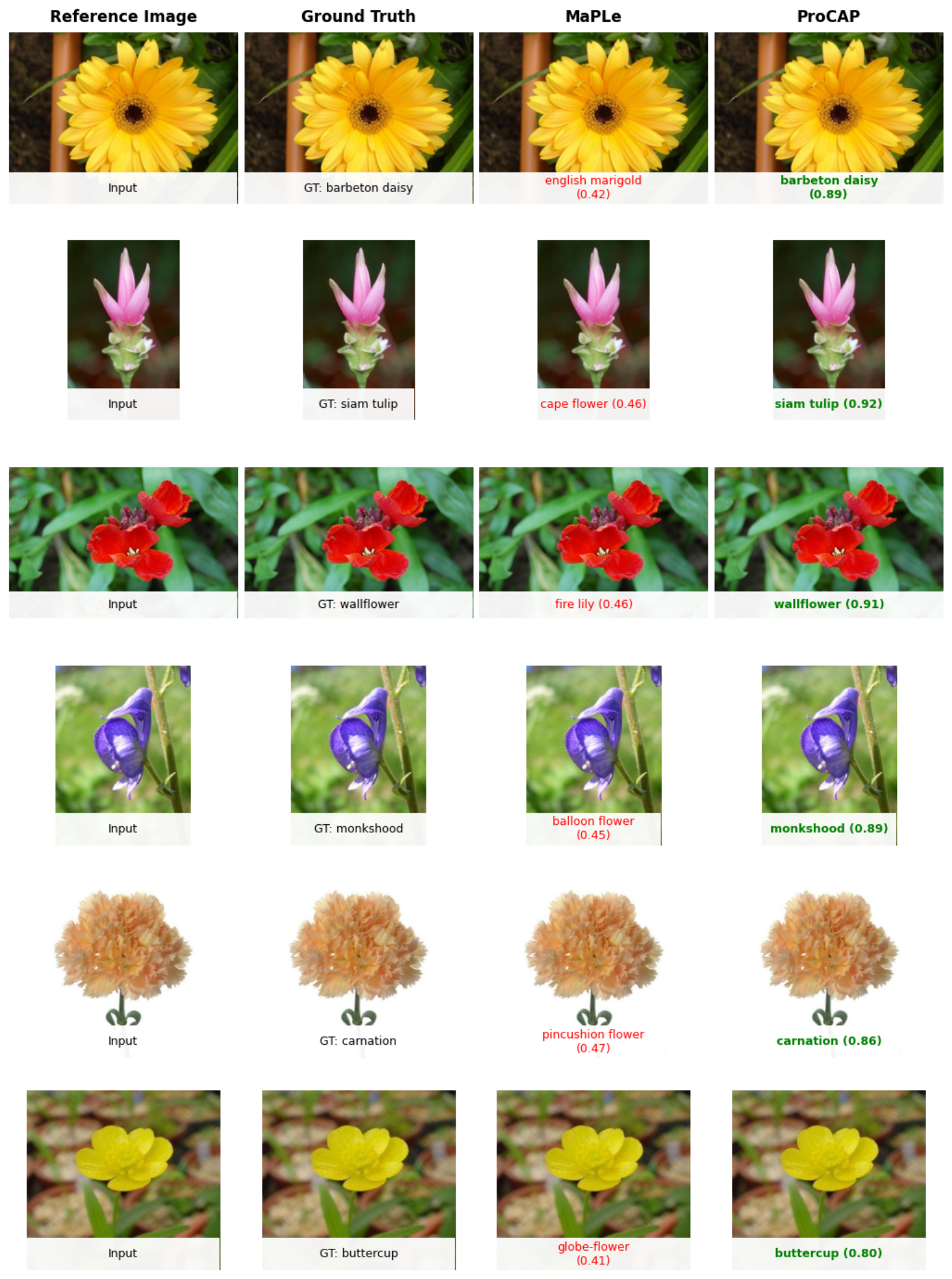}
  \caption{Qualitative prediction comparison of MaPLe and ProCAP on Oxford Flowers. Columns show the input, ground truth, MaPLe prediction, and ProCAP prediction; green/red denote correct/incorrect predictions.}
  \label{fig:qualitative_flowers_predictions}
\end{figure}

\begin{figure}[H]
  \centering
  \includegraphics[width=0.96\textwidth,height=0.39\textheight,keepaspectratio]{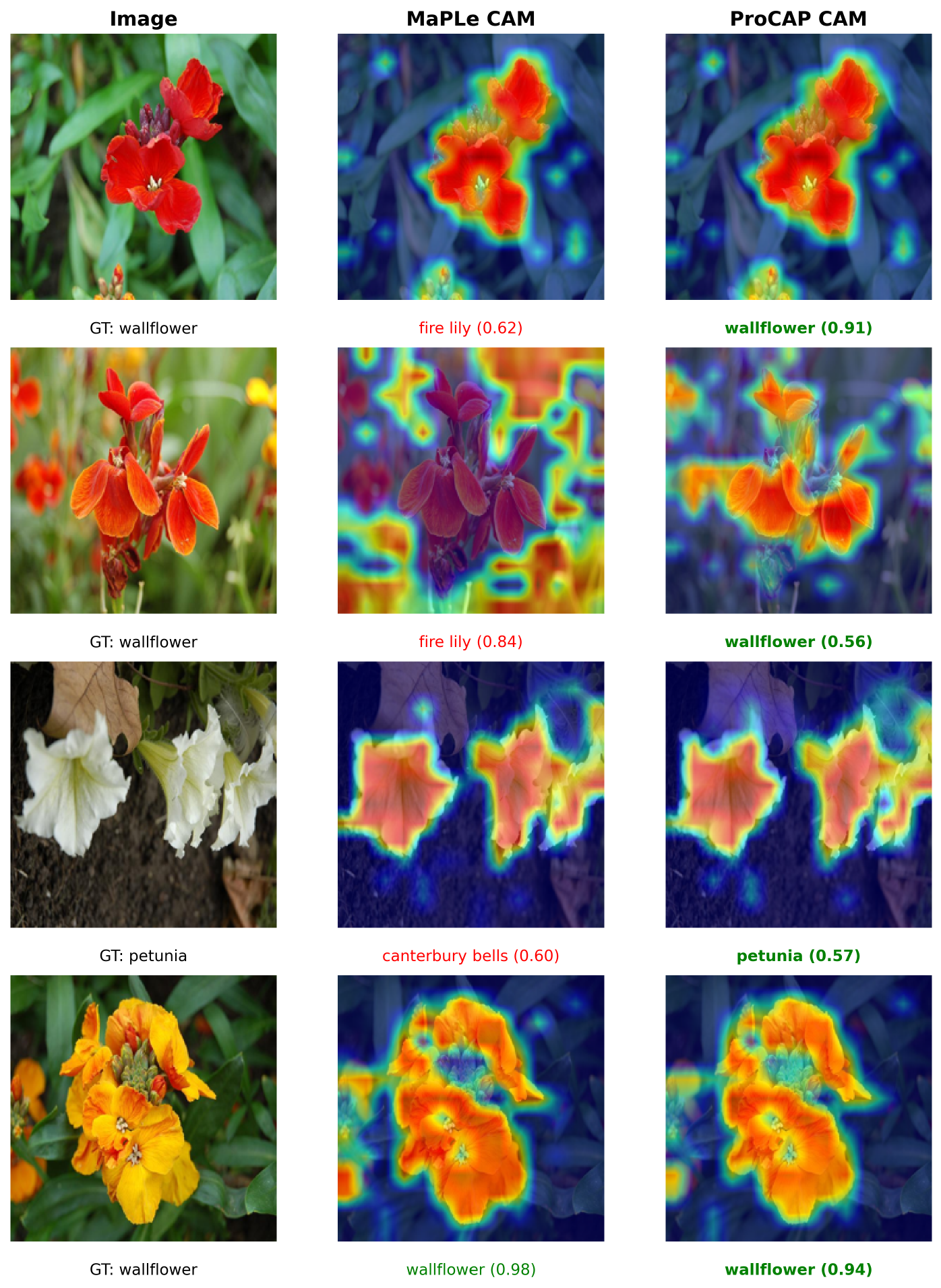}
  \caption{EigenCAM comparison of MaPLe and ProCAP on Oxford Flowers. Warmer regions indicate stronger contributions to the prediction.}
  \label{fig:qualitative_flowers_cam}
\end{figure}

\begin{figure}[H]
  \centering
  \includegraphics[width=0.82\textwidth]{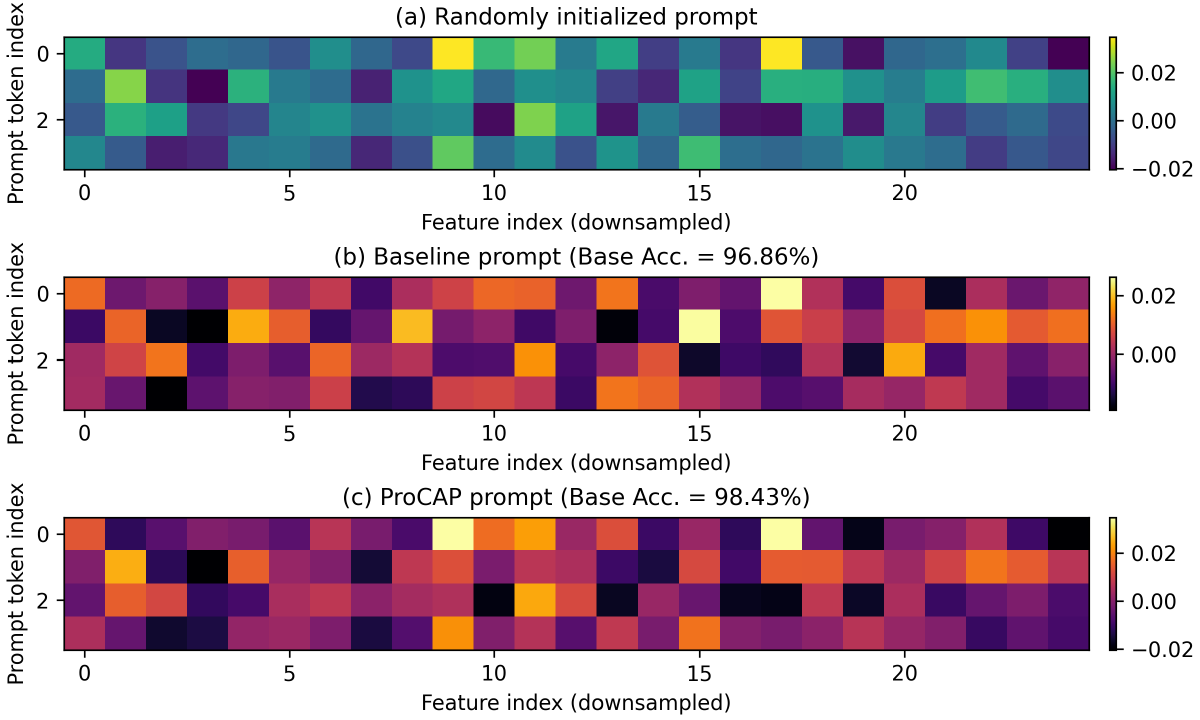}
  \caption{Learned text-prompt means ($\mu$) on EuroSAT base classes: (a) random initialization, (b) the trained baseline (96.86\% base accuracy), and (c) ProCAP (98.43\%).}
  \label{fig:prompt_vis}
\end{figure}

We provide three complementary qualitative analyses. Figure~\ref{fig:qualitative_flowers_predictions} compares final MaPLe~\cite{khattak2023maple} and ProCAP predictions on challenging Oxford Flowers~\cite{nilsback2008automated} examples; Figure~\ref{fig:qualitative_flowers_cam} compares EigenCAM visualizations~\cite{muhammad2020eigencam}; and Figure~\ref{fig:prompt_vis} visualizes the evolution of learned prompt tokens. The prediction and CAM views serve complementary purposes: the former shows whether the final class decision is corrected, while the latter shows where each model concentrates evidence. We use identical samples and CAM extraction/rendering settings for MaPLe and ProCAP so that the visual comparison reflects model behavior rather than differences in the visualization pipeline. Across the selected difficult fine-grained examples, ProCAP more often predicts the correct class and concentrates on discriminative flower regions, whereas MaPLe is more prone to confusion or diffuse attention toward less informative regions.

\newpage
\flushbottom

Figure~\ref{fig:prompt_vis} shows that training produces structured prompt patterns and that ProCAP learns a distinct pattern from the baseline.

\subsection{Stability Analysis}
Table~\ref{tab:procap_seed_stability_supp} reports three-seed variability across all 11 base-to-novel benchmarks.

\begin{table}[H]
\centering
\caption{Three-seed stability of ProCAP on base-to-novel generalization (mean $\pm$ standard deviation). The Average-row HM is the arithmetic mean of the 11 dataset-wise HMs; Table~\ref{tab:base_to_novel} reports aggregate HM from averaged Base and Novel accuracy.}
\label{tab:procap_seed_stability_supp}
\small
\renewcommand{\arraystretch}{1.08}
\begin{tabular*}{0.86\textwidth}{@{\extracolsep{\fill}}lccc@{}}
\toprule
Dataset & Base & Novel & \HM{} \\
\midrule
Average      & 85.43$\pm$0.18 & 77.61$\pm$0.31 & 81.12$\pm$0.24 \\
ImageNet     & 77.33$\pm$0.12 & 70.00$\pm$0.18 & 73.48$\pm$0.15 \\
Caltech101   & 98.97$\pm$0.10 & 95.31$\pm$0.21 & 97.11$\pm$0.15 \\
OxfordPets   & 96.49$\pm$0.17 & 98.21$\pm$0.12 & 97.34$\pm$0.14 \\
StanfordCars & 78.65$\pm$0.29 & 77.26$\pm$0.34 & 77.95$\pm$0.31 \\
Flowers102   & 98.39$\pm$0.22 & 75.60$\pm$0.71 & 85.50$\pm$0.49 \\
Food101      & 90.60$\pm$0.13 & 92.10$\pm$0.19 & 91.34$\pm$0.16 \\
FGVCAircraft & 45.92$\pm$0.46 & 36.77$\pm$0.58 & 40.84$\pm$0.51 \\
SUN397       & 81.90$\pm$0.25 & 79.45$\pm$0.32 & 80.66$\pm$0.28 \\
DTD          & 85.19$\pm$0.53 & 64.13$\pm$0.77 & 73.17$\pm$0.63 \\
EuroSAT      & 98.21$\pm$0.31 & 84.74$\pm$0.89 & 90.98$\pm$0.62 \\
UCF101       & 88.11$\pm$0.24 & 80.10$\pm$0.38 & 83.91$\pm$0.31 \\
\bottomrule
\end{tabular*}
\end{table}

\subsection{Runtime and Training Analysis}

We report runtime trade-offs and the effect of training duration to make the computational behavior of ProCAP explicit.

\begin{table}[H]
\centering
\caption{Training time, inference throughput, and accuracy on two representative datasets. All methods use CLIP ViT-B/16~\cite{radford2021clip} in the same 16-shot setting; runtime is measured over a fixed 10-epoch benchmark on one NVIDIA A100 GPU with AMP.}
\label{tab:eff_runtime_supp}
\small
\setlength{\tabcolsep}{4pt}
\renewcommand{\arraystretch}{1.03}
\begin{tabular}{l l r r r}
\toprule
Dataset & Method & Time (s) & FPS & Acc (\%) \\
\midrule
Flowers102 & MaPLe-style \tabcite{khattak2023maple} & 112.6 & 415.9 & 95.25 \\
Flowers102 & ProCAP (Full) & 169.1 & 278.8 & \best{97.39} \\
\midrule
UCF101 & MaPLe-style \tabcite{khattak2023maple} & 131.3 & 415.6 & 85.21 \\
UCF101 & ProCAP (Full) & 171.4 & 320.9 & \best{87.88} \\
\bottomrule
\end{tabular}
\end{table}

Table~\ref{tab:eff_runtime_supp} shows that ProCAP is slower during both training and inference, as expected from the added cross-modal interaction, while providing clear accuracy gains.

\begin{table}[H]
\centering
\caption{Effect of training duration on ProCAP for base-to-novel generalization, averaged over 11 datasets. HM is computed from the listed average Base and Novel accuracies using Eq.~\eqref{eq:hm}.}
\label{tab:procap_epochs_supp}
\small
\setlength{\tabcolsep}{6pt}
\renewcommand{\arraystretch}{1.08}
\begin{tabular}{l c c c c}
\toprule
Model & Epochs & Base & Novel & \HM{} \\
\midrule
ProCAP & 10 & 84.05 & 76.40 & 80.04 \\
ProCAP & 25 & 84.28 & 77.15 & 80.56 \\
ProCAP & 50 & \best{85.43} & \best{77.61} & \best{81.33} \\
\bottomrule
\end{tabular}
\end{table}

Longer training primarily improves base performance while preserving novel-class transfer; the best aggregate HM is obtained at 50 epochs.

\subsection{Implementation Details and Reproducibility}

Unless otherwise stated, the additional experiments use the same training and evaluation pipeline as the main paper.

\subsubsection{Training Setup}
All experiments use a frozen CLIP ViT-B/16 backbone~\cite{radford2021clip}; only prompt-learning parameters and ProCAP's additional learnable components are updated. Training uses AdamW with cosine scheduling and linear warmup for 50 epochs, batch size 32 (training) / 100 (evaluation), learning rate $2.5\times10^{-4}$, weight decay 0.03, warmup 3 epochs, and AMP. Main base-to-novel results average three random seeds; auxiliary experiments use seed 1 unless noted. Images are resized to $224\times224$, normalized with CLIP preprocessing, and trained with random resized crop and horizontal flip; EuroSAT additionally uses mild color jitter and small rotations.

The main configuration uses four context tokens (N\_CTX = 4), prompt depth nine (PROMPT\_\allowbreak DEPTH = 9), dataset-appropriate prompt initialization, and PREC = ``amp''. Multiple templates use the delimiter \texttt{||} only as a parser separator before tokenization.

\subsubsection{Evaluation Protocol}
For base-to-novel evaluation we report base accuracy, novel accuracy, and HM. For domain generalization, prompts are optimized on ImageNet and transferred directly to the target shift benchmarks without target supervision. Evaluation uses deterministic preprocessing only; horizontal-flip test-time augmentation is disabled (TTA\_HFLIP = False), and each test image is evaluated once.

For Table~\ref{tab:eff_runtime_supp}, all methods use the same environment, preprocessing, and evaluation conditions. The runtime numbers are therefore intended as a fair relative comparison rather than a hardware-independent deployment benchmark.

\subsubsection{Qualitative Visualization Protocol}
Figures~\ref{fig:qualitative_flowers_predictions} and~\ref{fig:qualitative_flowers_cam} use the same Oxford Flowers samples. EigenCAM~\cite{muhammad2020eigencam} extraction and rendering settings are identical for MaPLe and ProCAP so the comparison isolates model behavior rather than visualization differences.

\subsubsection{Implementation Environment}
All experiments are implemented in PyTorch and run on NVIDIA GPUs using the same codebase, backbone initialization, training framework, and evaluation pipeline as the main experiments.

\subsection{Dataset Details}

Table~\ref{tab:dataset_summary_supp} summarizes the evaluation datasets. Prompt-learning counts follow the fixed CoOp~\cite{zhou2022coop}/MaPLe benchmark splits and should be interpreted as benchmark split sizes rather than, where applicable, the size of the full raw collection.

\begin{table}[H]
\centering
\caption{Datasets and benchmark split sizes used in our experiments.}
\label{tab:dataset_summary_supp}
\scriptsize
\setlength{\tabcolsep}{3.5pt}
\renewcommand{\arraystretch}{0.90}
\resizebox{\textwidth}{!}{%
\begin{tabular}{lccccc}
\toprule
Dataset & Classes & Train & Val & Test & Description \\
\midrule
ImageNet & 1,000 & 1.28M & -- & 50,000 & Generic object recognition \\
Caltech101 & 100 & 4,128 & 1,649 & 2,465 & Generic object recognition \\
OxfordPets & 37 & 2,944 & 736 & 3,669 & Fine-grained pets \\
StanfordCars & 196 & 6,509 & 1,635 & 8,041 & Fine-grained cars \\
Flowers102 & 102 & 4,093 & 1,633 & 2,463 & Fine-grained flowers \\
Food101 & 101 & 50,500 & 20,200 & 30,300 & Fine-grained food \\
\shortstack[l]{FGVC\\Aircraft} & 100 & 3,334 & 3,333 & 3,333 & Fine-grained aircraft \\
SUN397 & 397 & 15,880 & 3,970 & 19,850 & Scene classification \\
DTD & 47 & 2,820 & 1,128 & 1,692 & Texture classification \\
EuroSAT & 10 & 13,500 & 5,400 & 8,100 & Satellite land-use classification \\
UCF101 & 101 & 7,639 & 1,898 & 3,783 & Action recognition \\
\midrule
ImageNetV2 & 1,000 & -- & -- & 10,000 & ImageNet distribution shift \\
ImageNet-Sketch & 1,000 & -- & -- & 50,889 & Sketch-style ImageNet images \\
ImageNet-A & 200 & -- & -- & 7,500 & Natural adversarial examples \\
ImageNet-R & 200 & -- & -- & 30,000 & Renditions of ImageNet classes \\
\bottomrule
\end{tabular}}
\end{table}

\subsection{Prompt Templates}
We did not exhaustively evaluate template combinations; systematic template sensitivity remains future work.

\begin{table}[H]
\centering
\caption{Representative prompt templates used for each dataset. The placeholder \texttt{\{\}} is replaced by the class name before tokenization.}
\label{tab:templates_supp}
\footnotesize
\setlength{\tabcolsep}{4pt}
\renewcommand{\arraystretch}{1.10}
\begin{tabular}{>{\raggedright\arraybackslash}p{2.15cm} >{\raggedright\arraybackslash}p{5.9cm} >{\raggedright\arraybackslash}p{5.9cm}}
\toprule
Dataset & Template 1 & Template 2 \\
\midrule
ImageNet & a photo of a \{\} & a close-up photo of a \{\} \\
Caltech101 & a photo of a \{\} & an image of a \{\} \\
OxfordPets & a photo of a \{\} & a close-up photo of a \{\} \\
StanfordCars & a photo of a \{\} car & a close-up photo of a \{\} car \\
Flowers102 & a photo of a \{\} flower & a close-up photo of a \{\} flower \\
Food101 & a photo of \{\} & a close-up photo of \{\} \\
FGVCAircraft & a photo of an \{\} aircraft & a photo of an \{\} airplane \\
SUN397 & a photo of a \{\} scene & an indoor scene of a \{\} \\
DTD & a photo of a \{\} texture & a close-up photo of a \{\} texture \\
EuroSAT & a centered satellite photo of a \{\} & a satellite image of a \{\} \\
UCF101 & a video of a person doing \{\} & a video frame of a person doing \{\} \\
\bottomrule
\end{tabular}
\end{table}

The table lists two representative configured templates per dataset. When more than one template is specified in the configuration, \texttt{||} is used only as an internal parser delimiter and is removed before tokenization; it is not part of the text presented to CLIP.

%% file: main.bib
@inproceedings{radford2021clip,
  title     = {Learning Transferable Visual Models From Natural Language Supervision},
  author    = {Alec Radford and Jong Wook Kim and Chris Hallacy and Aditya Ramesh and Gabriel Goh and Sandhini Agarwal and Girish Sastry and Amanda Askell and Pamela Mishkin and Jack Clark and Gretchen Krueger and Ilya Sutskever},
  booktitle = {Proceedings of the 38th International Conference on Machine Learning (ICML)},
  series    = {Proceedings of Machine Learning Research},
  volume    = {139},
  pages     = {8748--8763},
  year      = {2021}
}

@article{zhou2022coop,
  title   = {Learning to Prompt for Vision-Language Models},
  author  = {Kaiyang Zhou and Jingkang Yang and Chen Change Loy and Ziwei Liu},
  journal = {International Journal of Computer Vision},
  volume  = {130},
  number  = {9},
  pages   = {2337--2348},
  year    = {2022},
  doi     = {10.1007/s11263-022-01653-1}
}

@inproceedings{zhou2022cocoop,
  title     = {Conditional Prompt Learning for Vision-Language Models},
  author    = {Kaiyang Zhou and Jingkang Yang and Chen Change Loy and Ziwei Liu},
  booktitle = {Proceedings of the IEEE/CVF Conference on Computer Vision and Pattern Recognition (CVPR)},
  year      = {2022}
}

@inproceedings{khattak2023maple,
  title     = {MaPLe: Multi-modal Prompt Learning},
  author    = {Muhammad Uzair Khattak and Hanoona Abdul Rasheed and Muhammad Maaz and Salman Khan and Fahad Shahbaz Khan},
  booktitle = {Proceedings of the IEEE/CVF Conference on Computer Vision and Pattern Recognition (CVPR)},
  year      = {2023}
}

@article{wu2023ammpl,
  title   = {Adaptive Multi-Modality Prompt Learning},
  author  = {Zongqian Wu and Yujing Liu and Mengmeng Zhan and Jialie Shen and Ping Hu and Xiaofeng Zhu},
  journal = {arXiv preprint arXiv:2312.00823},
  year    = {2023},
  doi     = {10.48550/arXiv.2312.00823}
}

@inproceedings{khattak2023selfreg,
  title     = {Self-Regulating Prompts: Foundational Model Adaptation Without Forgetting},
  author    = {Muhammad Uzair Khattak and Syed Talal Wasim and Muzammal Naseer and Salman Khan and Ming-Hsuan Yang and Fahad Shahbaz Khan},
  booktitle = {Proceedings of the IEEE/CVF International Conference on Computer Vision (ICCV)},
  year      = {2023}
}

@inproceedings{kim2024aapl,
  title     = {AAPL: Adding Attributes to Prompt Learning for Vision-Language Models},
  author    = {Gahyeon Kim and Sohee Kim and Seokju Lee},
  booktitle = {Proceedings of the IEEE/CVF Conference on Computer Vision and Pattern Recognition Workshops (CVPRW)},
  year      = {2024},
  eprint    = {2404.16804},
  archivePrefix = {arXiv},
  primaryClass  = {cs.CV}
}

@inproceedings{roy2024coprompt,
  title     = {Consistency-guided Prompt Learning for Vision-Language Models},
  author    = {Shuvendu Roy and Ali Etemad},
  booktitle = {International Conference on Learning Representations (ICLR)},
  year      = {2024},
  url       = {https://openreview.net/forum?id=wsRXwlwx4w}
}

@article{du2024mocoop,
  title   = {Mixture of Prompt Learning for Vision Language Models},
  author  = {Yu Du and Tong Niu and Rong Zhao},
  journal = {arXiv preprint arXiv:2409.12011},
  year    = {2024}
}

@inproceedings{wu2024caspl,
  title     = {Cascade Prompt Learning for Vision-Language Model Adaptation},
  author    = {Ge Wu and Xin Zhang and Zheng Li and Zhaowei Chen and Jiajun Liang and Jian Yang and Xiang Li},
  booktitle = {Proceedings of the European Conference on Computer Vision (ECCV)},
  year      = {2024},
  eprint    = {2409.17805},
  archivePrefix = {arXiv},
  primaryClass  = {cs.CV}
}

@inproceedings{zhang2024dept,
  title     = {DePT: Decoupled Prompt Tuning},
  author    = {Ji Zhang and Shihan Wu and Lianli Gao and Heng Tao Shen and Jingkuan Song},
  booktitle = {Proceedings of the IEEE/CVF Conference on Computer Vision and Pattern Recognition (CVPR)},
  year      = {2024},
  eprint    = {2309.07439},
  archivePrefix = {arXiv},
  primaryClass  = {cs.CV}
}

@inproceedings{chen2024mma,
  title     = {MMA: Multi-Modal Adapter for Vision-Language Models},
  author    = {Lingxiao Yang and Ru-Yuan Zhang and Yanchen Wang and Xiaohua Xie},
  booktitle = {Proceedings of the IEEE/CVF Conference on Computer Vision and Pattern Recognition (CVPR)},
  year      = {2024},
  pages     = {23826--23837}
}

@inproceedings{du2024ipo,
  title     = {IPO: Interpretable Prompt Optimization for Vision-Language Models},
  author    = {Yingjun Du and Wenfang Sun and Cees G. M. Snoek},
  booktitle = {Advances in Neural Information Processing Systems (NeurIPS)},
  year      = {2024},
  eprint    = {2410.15397},
  archivePrefix = {arXiv},
  primaryClass  = {cs.CV}
}

@inproceedings{alipour2024stylepro,
  title     = {Style-Pro: Style-Guided Prompt Learning for Generalizable Vision-Language Models},
  author    = {Niloufar Alipour Talemi and Hossein Kashiani and Fatemeh Afghah},
  booktitle = {Proceedings of the IEEE/CVF Winter Conference on Applications of Computer Vision (WACV)},
  year      = {2025},
  eprint    = {2411.16018},
  archivePrefix = {arXiv},
  primaryClass  = {cs.CV}
}

@inproceedings{li2025dpc,
  title     = {DPC: Dual-Prompt Collaboration for Tuning Vision-Language Models},
  author    = {Haoyang Li and Liang Wang and Chao Wang and Jing Jiang and Yan Peng and Guodong Long},
  booktitle = {Proceedings of the IEEE/CVF Conference on Computer Vision and Pattern Recognition (CVPR)},
  month     = {June},
  year      = {2025},
  pages     = {25623--25632}
}

@inproceedings{cheng2025vamp,
  title     = {VaMP: Variational Multi-Modal Prompt Learning for Vision-Language Models},
  author    = {Silin Cheng and Kai Han},
  booktitle = {Advances in Neural Information Processing Systems (NeurIPS)},
  year      = {2025},
  eprint    = {2511.22664},
  archivePrefix = {arXiv},
  primaryClass  = {cs.CV}
}

@article{yang2025mugcp,
  title   = {Multi-modal Mutual-Guidance Conditional Prompt Learning for Vision-Language Models},
  author  = {Shijun Yang and Xiang Zhang and Wanqing Zhao and Hangzai Luo and Sheng Zhong and Jinye Peng and Jianping Fan},
  journal = {arXiv preprint arXiv:2507.08410},
  year    = {2025},
  url     = {https://arxiv.org/abs/2507.08410}
}

@article{yin2024incpl,
  title   = {In-context Prompt Learning for Test-time Vision Recognition with Frozen Vision-Language Model},
  author  = {Junhui Yin and Xinyu Zhang and Lin Wu and Xiaojie Wang},
  journal = {arXiv preprint arXiv:2403.06126},
  year    = {2024},
  doi     = {10.48550/arXiv.2403.06126}
}

@article{dai2024muap,
  title   = {MuAP: Multi-step Adaptive Prompt Learning for Vision-Language Model with Missing Modality},
  author  = {Ruiting Dai and Yuqiao Tan and Lisi Mo and Tao He and Ke Qin and Shuang Liang},
  journal = {arXiv preprint arXiv:2409.04693},
  year    = {2024}
}

@inproceedings{mistretta2024kdpl,
  title     = {Improving Zero-shot Generalization of Learned Prompts via Unsupervised Knowledge Distillation},
  author    = {Marco Mistretta and Alberto Baldrati and Marco Bertini and Andrew D. Bagdanov},
  booktitle = {Proceedings of the European Conference on Computer Vision (ECCV)},
  year      = {2024},
  eprint    = {2407.03056},
  archivePrefix = {arXiv},
  primaryClass  = {cs.CV}
}

@inproceedings{sptr2025aaai,
  title     = {A Similarity Paradigm Through Textual Regularization Without Forgetting},
  author    = {Fangming Cui and Jan Fong and Rongfei Zeng and Xinmei Tian and Jun Yu},
  booktitle = {Proceedings of the AAAI Conference on Artificial Intelligence},
  year      = {2025},
  eprint    = {2502.14376},
  archivePrefix = {arXiv},
  primaryClass  = {cs.CV}
}

@inproceedings{recht2019imagenetv2,
  title     = {Do ImageNet Classifiers Generalize to ImageNet?},
  author    = {Benjamin Recht and Rebecca Roelofs and Ludwig Schmidt and Vaishaal Shankar},
  booktitle = {Proceedings of the 36th International Conference on Machine Learning (ICML)},
  year      = {2019},
  eprint    = {1902.10811},
  archivePrefix = {arXiv}
}

@inproceedings{wang2019learning,
  title     = {Learning Robust Global Representations by Penalizing Local Predictive Power},
  author    = {Haohan Wang and Songwei Ge and Eric P. Xing and Zachary C. Lipton},
  booktitle = {Advances in Neural Information Processing Systems (NeurIPS)},
  year      = {2019},
  eprint    = {1905.13549},
  archivePrefix = {arXiv}
}

@inproceedings{hendrycks2019natural,
  title     = {Natural Adversarial Examples},
  author    = {Dan Hendrycks and Kevin Zhao and Steven Basart and Jacob Steinhardt and Dawn Song},
  booktitle = {Proceedings of the IEEE/CVF Conference on Computer Vision and Pattern Recognition (CVPR)},
  year      = {2021},
  pages     = {15262--15271}
}

@inproceedings{rahman2025dimple,
  title     = {DiMPLe - Disentangled Multi-Modal Prompt Learning: Enhancing Out-Of-Distribution Alignment with Invariant and Spurious Feature Separation},
  author    = {Rahman, Umaima and Yaqub, Mohammad and Mahapatra, Dwarikanath},
  booktitle = {Proceedings of the IEEE/CVF International Conference on Computer Vision (ICCV)},
  month     = oct,
  year      = {2025},
  pages     = {1634--1643}
}

@inproceedings{yao2023kgcoop,
  author    = {Hantao Yao and Rui Zhang and Changsheng Xu},
  title     = {Visual-Language Prompt Tuning With Knowledge-Guided Context Optimization},
  booktitle = {Proceedings of the IEEE/CVF Conference on Computer Vision and Pattern Recognition (CVPR)},
  month     = {June},
  year      = {2023},
  pages     = {6757--6767},
  url       = {https://openaccess.thecvf.com/content/CVPR2023/html/Yao_Visual-Language_Prompt_Tuning_With_Knowledge-Guided_Context_Optimization_CVPR_2023_paper.html}
}

@inproceedings{yang2024tcp,
  author    = {Hantao Yao and Rui Zhang and Changsheng Xu},
  title     = {TCP:Textual-based Class-aware Prompt tuning for Visual-Language Model},
  booktitle = {Proceedings of the IEEE/CVF Conference on Computer Vision and Pattern Recognition (CVPR)},
  month     = {June},
  year      = {2024},
  pages     = {23438--23448},
  eprint    = {2311.18231},
  archivePrefix = {arXiv},
  primaryClass  = {cs.CV},
  url       = {https://openaccess.thecvf.com/content/CVPR2024/html/Yao_TCPTextual-based_Class-aware_Prompt_tuning_for_Visual-Language_Model_CVPR_2024_paper.html}
}

@inproceedings{farina2025two,
  author    = {Matteo Farina and Massimiliano Mancini and Giovanni Iacca and Elisa Ricci},
  title     = {Rethinking Few-Shot Adaptation of Vision-Language Models in Two Stages},
  booktitle = {Proceedings of the IEEE/CVF Conference on Computer Vision and Pattern Recognition (CVPR)},
  month     = {June},
  year      = {2025},
  pages     = {29989--29998},
  eprint    = {2503.11609},
  archivePrefix = {arXiv},
  primaryClass  = {cs.CV},
  url       = {https://openaccess.thecvf.com/content/CVPR2025/html/Farina_Rethinking_Few-Shot_Adaptation_of_Vision-Language_Models_in_Two_Stages_CVPR_2025_paper.html}
}

@inproceedings{zhu2025skiptuning,
  author    = {Shihan Wu and Ji Zhang and Pengpeng Zeng and Lianli Gao and Jingkuan Song and Heng Tao Shen},
  title     = {Skip Tuning: Pre-trained Vision-Language Models are Effective and Efficient Adapters Themselves},
  booktitle = {Proceedings of the IEEE/CVF Conference on Computer Vision and Pattern Recognition (CVPR)},
  month     = {June},
  year      = {2025},
  pages     = {14723--14732},
  eprint    = {2412.11509},
  archivePrefix = {arXiv},
  primaryClass  = {cs.CV},
  url       = {https://openaccess.thecvf.com/content/CVPR2025/html/Wu_Skip_Tuning_Pre-trained_Vision-Language_Models_are_Effective_and_Efficient_Adapters_CVPR_2025_paper.html}
}

@inproceedings{guo2025mmrl,
  author    = {Yuncheng Guo and Xiaodong Gu},
  title     = {MMRL: Multi-Modal Representation Learning for Vision-Language Models},
  booktitle = {Proceedings of the IEEE/CVF Conference on Computer Vision and Pattern Recognition (CVPR)},
  month     = {June},
  year      = {2025},
  pages     = {25015--25025},
  eprint    = {2503.08497},
  archivePrefix = {arXiv},
  primaryClass  = {cs.CV},
  url       = {https://openaccess.thecvf.com/content/CVPR2025/html/Guo_MMRL_Multi-Modal_Representation_Learning_for_Vision-Language_Models_CVPR_2025_paper.html}
}

@article{sparsekgcoop2025,
  author  = {Qiangxing Tian and Min Zhang},
  title   = {Enhancing Visual-Language Prompt Tuning Through Sparse Knowledge-Guided Context Optimization},
  journal = {Entropy},
  year    = {2025},
  volume  = {27},
  number  = {3},
  pages   = {301},
  doi     = {10.3390/e27030301},
  url     = {https://www.mdpi.com/1099-4300/27/3/301}
}

@inproceedings{hendrycks2021many,
  author    = {Dan Hendrycks and Steven Basart and Norman Mu and Saurav Kadavath and Frank Wang and Evan Dorundo and Rahul Desai and Tyler Zhu and Samyak Parajuli and Mike Guo and Dawn Song and Jacob Steinhardt and Justin Gilmer},
  title     = {The Many Faces of Robustness: A Critical Analysis of Out-of-Distribution Generalization},
  booktitle = {Proceedings of the IEEE/CVF International Conference on Computer Vision (ICCV)},
  month     = {October},
  year      = {2021},
  pages     = {8340--8349},
  url       = {https://openaccess.thecvf.com/content/ICCV2021/html/Hendrycks_The_Many_Faces_of_Robustness_A_Critical_Analysis_of_Out-of-Distribution_ICCV_2021_paper.html}
}

@inproceedings{muhammad2020eigencam,
  title     = {Eigen-CAM: Class Activation Map using Principal Components},
  author    = {Mohammed Bany Muhammad and Mohammed Yeasin},
  booktitle = {2020 International Joint Conference on Neural Networks (IJCNN)},
  pages     = {1--7},
  year      = {2020},
  doi       = {10.1109/IJCNN48605.2020.9206626}
}

@inproceedings{nilsback2008automated,
  title     = {Automated Flower Classification over a Large Number of Classes},
  author    = {Maria-Elena Nilsback and Andrew Zisserman},
  booktitle = {Indian Conference on Computer Vision, Graphics and Image Processing (ICVGIP)},
  year      = {2008}
}

@article{dong2026hi2ma,
  author  = {Chunru Dong and Tao Zhang and Feng Zhang and Qiang Hua and Jie Zhu and Yong Zhang},
  title   = {Hierarchical Intra-Inter Modal Adaptation for Vision-Language Models},
  journal = {Pattern Recognition},
  volume  = {180},
  pages   = {114200},
  year    = {2026},
  doi     = {10.1016/j.patcog.2026.114200},
  url     = {https://doi.org/10.1016/j.patcog.2026.114200}
}

@article{mei2026dcpl,
  author  = {Xiaoyong Mei and Chong Tang and Zhengqun Dai and Fudan Zheng and Kongwen Zhang and Tianyu Lin},
  title   = {Diversified Composite Prompting to Enhance Generalisation of Vision-Language Models},
  journal = {CAAI Transactions on Intelligence Technology},
  volume  = {11},
  number  = {3},
  pages   = {935--950},
  year    = {2026},
  doi     = {10.1049/cit2.70143},
  url     = {https://doi.org/10.1049/cit2.70143}
}

@article{zhang2026mpcle,
  author  = {Geyuan Zhang and Xiaofei Zhou and Gaopeng Gou and Gang Xiong and Li Guo},
  title   = {Multi-Modal Prompt Codebook Learning: Achieving Adaptive and Generalizable Prompting for {CLIP}-Based Visual Recognition},
  journal = {Information Sciences},
  volume  = {756},
  pages   = {123849},
  year    = {2026},
  doi     = {10.1016/j.ins.2026.123849},
  url     = {https://doi.org/10.1016/j.ins.2026.123849}
}

@inproceedings{ghiasvand2026mmlop,
  author    = {Sajjad Ghiasvand and Haniyeh Ehsani Oskouie and Mahnoosh Alizadeh and Ramtin Pedarsani},
  title     = {{MMLoP}: Multi-Modal Low-Rank Prompting for Efficient Vision-Language Adaptation},
  booktitle = {European Conference on Computer Vision (ECCV)},
  year      = {2026},
  note      = {Accepted paper; arXiv:2602.21397},
  doi       = {10.48550/arXiv.2602.21397},
  url       = {https://arxiv.org/abs/2602.21397}
}
